\documentclass{article}

\usepackage{microtype}
\usepackage{graphicx}
\usepackage{subfigure}
\usepackage{booktabs} % for professional tables
\usepackage{paralist} % for compactenum
\usepackage[T1]{fontenc}
\usepackage{xurl}

\usepackage{hyperref}

\usepackage[accepted]{icml2024}

\usepackage{amsmath}
\usepackage{amssymb}
\usepackage{mathtools}
\usepackage{amsthm}

\usepackage[capitalize,noabbrev]{cleveref}

\theoremstyle{plain}

\theoremstyle{definition}

\theoremstyle{remark}

\usepackage[textsize=tiny]{todonotes}

\icmltitlerunning{Explorations of Self-Repair in Language Models}

\begin{document}

\twocolumn[
\icmltitle{Explorations of Self-Repair in Language Models}

\begin{icmlauthorlist}
\icmlauthor{Cody Rushing}{UT}
\icmlauthor{Neel Nanda}{}

\end{icmlauthorlist}

\icmlaffiliation{UT}{University of Texas at Austin}

\icmlcorrespondingauthor{Cody Rushing}{thisiscodyr@gmail.com}
%\icmlcorrespondingauthor{Firstname2 Lastname2}{first2.last2@www.uk}

\icmlkeywords{Machine Learning, Mechanistic Interpretability, Self Repair, Ablations, Interpretability, Large Language Models}

%\vskip 0.3in
]



\printAffiliationsAndNotice{}  % leave blank if no need to mention equal contribution
\newcommand{\LN}{30\% }
\begin{abstract}
Prior interpretability research studying narrow distributions has preliminarily identified self-repair, a phenomena where if components in large language models are ablated, later components will change their behavior to compensate. Our work builds off this past literature, demonstrating that self-repair exists on a variety of models families and sizes when ablating individual attention heads on the full training distribution. We further show that on the full training distribution self-repair is imperfect, as the original direct effect of the head is not fully restored, and noisy, since the degree of self-repair varies significantly across different prompts (sometimes overcorrecting beyond the original effect). We highlight two different mechanisms that contribute to self-repair, including changes in the final LayerNorm scaling factor and sparse sets of neurons implementing Anti-Erasure. We additionally discuss the implications of these results for interpretability practitioners and close with a more speculative discussion on the mystery of why self-repair occurs in these models at all, highlighting evidence for the Iterative Inference hypothesis in language models, a framework that predicts self-repair. 

\end{abstract}
\begin{figure}[ht]
%%\vskip 0.2in
\begin{center}
\centerline{\includegraphics[width=\columnwidth]{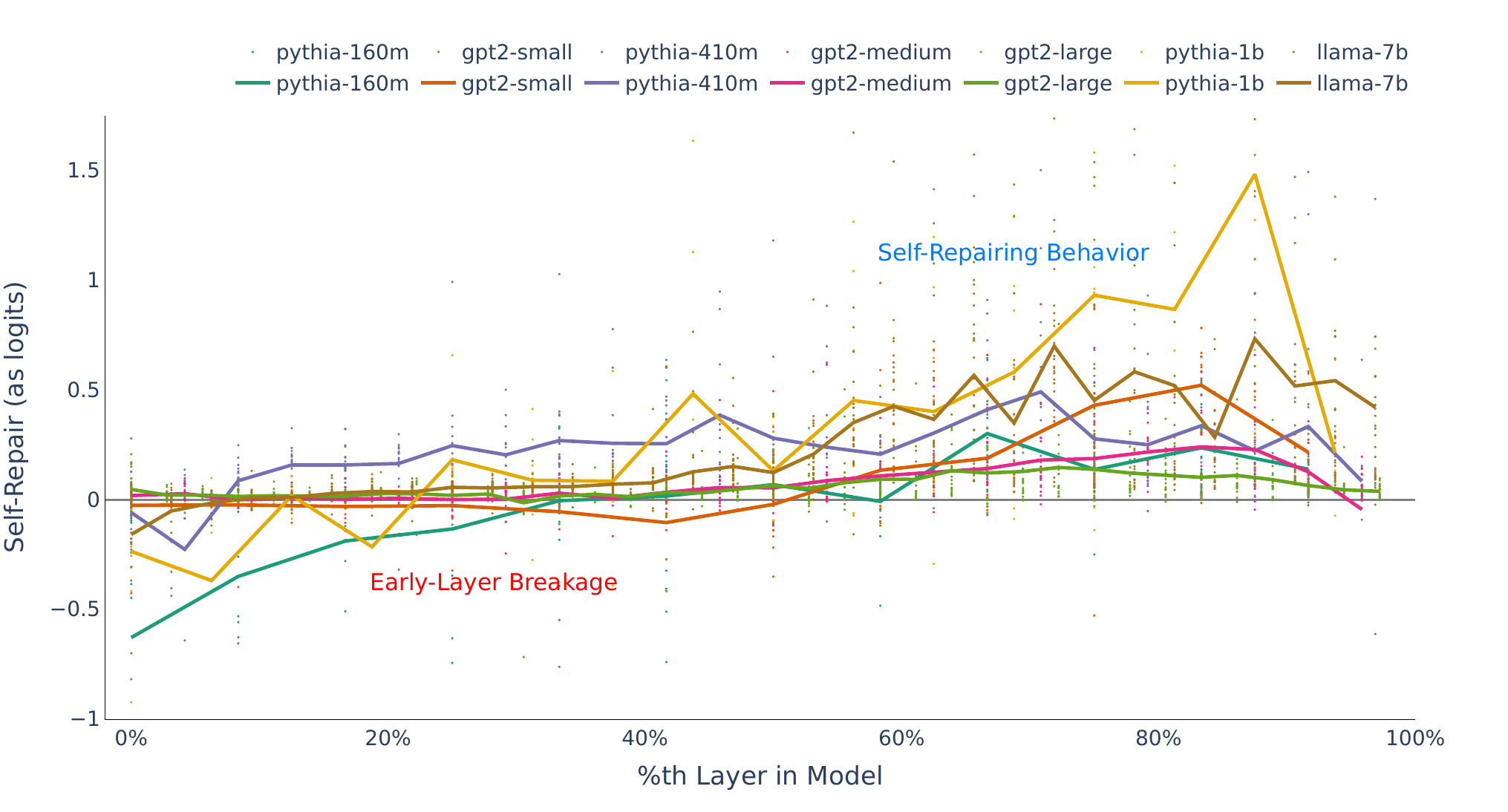}}%
\caption{We measure the self-repair of an attention head when resample ablated on the top 2\% of tokens according to its direct effect. For each model, we plot both the self-repair of the individual heads and a trend line that averages across the heads in each layer. Self-Repair exists across many later layers in different models, although the amount varies between heads.}
\label{fig:self-repair-significance}
\end{center}
\vskip -0.4in
\end{figure}

\section{Introduction}
\label{Introduction}
Interpretability efforts aim to reveal human-understandable behaviors in large language model components. A common interpretability technique is model ablations, where outputs of individual model components are replaced with outputs from other distributions. Ideally, if an attention head is critical for a task, ablating it would harm model performance on that task.

However, recent work \citep{mcgrath2023hydra, wang2023interpretability} has uncovered preliminary evidence of self-repair, a phenomena in large language models where components downstream of ablations compensate for them. As such, the ablation of model components doesn't always lead to easily predictable changes in model performance; instead, the removal of the components behavior can be compensated for by downstream components in a way which masks the loss of the original component. 

This is a challenge for interpretability efforts which rely on ablation-based metrics to define the importance of model components. In particular, self-repair mechanisms minimize the impact of ablating components deemed critical by other metrics.

%  This is often not the case, however, as 

Past literature has looked at self-repair in incomplete settings: self-repair was first discovered in the Indirect Object Identification distribution as "Backup Behavior" \citep{wang2023interpretability}, for which the behavior was explained partially as a byproduct of Copy Suppression \citep{mcdougall2023copy}. The self-repair phenomena was then further explored in  \citealt{mcgrath2023hydra}, but only across the Counterfact dataset and with the ablation of entire model layers.

We strengthen this prior work by investigating self-repair across the whole pretraining distribution, by focusing on individual attention heads, and by investigating the mechanisms behind self-repair on the whole distribution. Our key findings are:

%We focus on self-repair of direct effects, rather than self-repair of composition between pairs of components. 

\begin{compactenum}

\item Direct effect self-repair is an imperfect, noisy process. It occurs across the full pretraining distribution, even while ablating individual heads (rather than full layers).
\item  A nontrivial fraction - possibly \LN - of the self-repairing of direct effects can be attributed to just the effect of ablations on the LayerNorm normalization factor.
\item  MLP Erasure, a mechanism in which MLP layers `erase' outputs from earlier model components and is known to explain self-repair \citep{mcgrath2023hydra}, is powered by a sparse set of Erasure neurons. The exact neurons vary between prompts.
\end{compactenum}

We end with some discussion on the implications of these results for interpretability research and some present preliminary evidence for the Iterative Inference hypothesis, a speculative framework that predicts self-repair as a side effect. All the code for the experiments used in this paper is provided at \url{https://github.com/starship006/backup_research}.

\section{Self-Repair on the Full Distribution Exists, but is Incomplete and Noisy}

In this section, we confirm that across models, individual attention heads are `self-repaired' on a general pretraining distribution. We detail how we measure self-repair, demonstrate how individual attention heads are imperfectly self-repaired across the entire pretraining distribution, and highlight how it is noisy.

\subsection{Defining Self-Repair}
\label{self-repair-metric}
We first detail how we define direct effect and self-repair. Our methodology aligns with prior research on self-repair, although not mirroring it identically (we detail how our work differs from prior works in Section \ref{sec:related-works}). Let $(x_1, x_2, ..., x_{n-1}, x_n)$ be a sequence of $n$ tokens we pass into a language model. A language model maps each contiguous prefix of input tokens $(x_1, x_2, ..., x_{k-1}, x_k)$ to a final activation $r_k$, ultimately producing $n$ final activations $(r_1, r_2, ..., r_{n-1}, r_n)$ for each prefix. 

We follow \citealt{elhage2021mathematical}  in calling these the final residual streams (as they are the output of the final MLP layer plus the final skip connection). An important fact about residual streams is that each final residual stream $r_k$ can be decomposed into the sum of the output of each layer, plus the original embeddings \citep{elhage2021mathematical}. We refer to anything which adds its output into the residual stream as a model component - components can include attention layers, individual attention heads, MLP layers, or the individual neurons within them (for simplicity, we also count the model embeddings as a component). Formally, for a decomposition into components $C^i$, define $C^i_k$ as the output on position $k$, $r_k = \sum_i C^i_k$. 

For a given final residual stream activation $r_k$, the model produces a distribution of logits $l_k$, obtained from then applying a normalization function $LN$, such as LayerNorm or RMSNorm, to the residual stream corresponding to $k$-th position, $res \in R^{d_{model}}$ (where $d_{model}$ is the dimensionality of the model's hidden states), and multiplying it by the unembedding matrix $W_U \in \mathbb{R}^{V \times d_{model}}$ of the model: $l_k = W_U \cdot LN(r_k)$. We center the logits so the average logit is zero, which prevents `identical translations in logits' from occuring (see Appendix \ref{center-logits}).

$W_U$ is a linear map and we can take a linear approximation to $LN$ on a given input  \cite{elhage2021mathematical}. The composition of two linear functions is a linear function, i.e. a matrix multiplication plus a bias, so $l_k = W r_k + b$ for some $W$ and $b$ (which vary with the input). Because $r_k$ is the sum of component outputs, $l_k = W (\sum_i C^i_k) + b = (\sum_i W C^i_k) + b$. 

We often just care about the logit of the correct next token $x_{k+1}$, which corresponds to a single element of $l_k$. This is $logit_{clean} = W[x_{k+1}]^T \cdot r_k + b[x_{k+1}]$, where $W[x_{k+1}]$ is the $x_{k+1}$-th row of $W$ and $b[x_{k+1}]$ is the $x_{k+1}$-th element of $b$ (which are a $\mathbb{R}^{d_{model}}$ vector and a scalar).

Using the decomposition of $l_k$, we can define $W[x_{k+1}]^T \cdot C^i_k$ as the \textbf{direct effect} of component $i$ on position $k$. Intuitively, a component can help a model predict the final answer by either producing intermediate representations that are used by later components, or by directly boosting the correct next token: the direct effect captures the latter effect only, measuring how much a component's output $C_k^i$ writes its output in the direction corresponding to the next correct token, $W[x_{k+1}]$.

Notice that just as the residual stream can be mostly decomposed into the sum of outputs of all model components, the logit of the correct next token $logit_{clean}$ can be mostly decomposed into the sum of the direct effects of all model components. Ideally, if a component has a high direct effect, it was likely `important' for predicting the next token.

We contrast the direct effect with ablation-based metrics. Intuitively, ablation-based metrics measure the full effect of how a component helps the model produce the output, both via producing intermediate representations used by later components and the direct effect. We perform resample ablations, a technique that replaces the output of an head as if it were from a different set of pretraining tokens \citep{chan2022causal}. We run the model again on a different set of input tokens $(x'_1, x'_2, ..., x'_{n-1}, x'_n)$ and store the new output of an attention head $C'_{out}$. Then, when running the model on the original set of tokens $(x_1, x_2, ..., x_{n-1}, x_n)$, we intervene during the forward pass to replace the original output of the head $C_{out}$ with the new output $C'_{out}$, and then continue the forward pass, a technique known as activation patching or causal mediation analysis \citep{vig2020causal}.

Let the new logit distribution under this ablation be $l'_{k}$ and the new logit of the correct next token be $logit_{ablated}$. Notice that we can also compute the ablated direct effect of the attention head. If the output of an attention head has no `indirect' downstream effects—i.e., no downstream model component depends on the output of the attention head—then the change in the correct logit, $\Delta logit = logit_{ablated} - logit_{clean}$, would equal the change in the direct effect of the attention head, $\Delta DE_{head}$. However, in practice, these are rarely equal, and often $|\Delta DE_{head}| > |\Delta logit|$. This discrepancy is caused by later components changing their direct effects in a way that compensates for the resample ablated component, the phenomena we refer to as self-repair. We measure the `self-repair' that occurs as:
\begin{equation}
\label{eq:self-repair-eq}
\text{self-repair} = \Delta logit - \Delta DE_{head}
\end{equation}

When we resample ablate a head, the new activation comes from a prompt drawn from the pre-training distribution, and so is likely to be unrelated. The new direct effect when resample ablating an arbitrary head is near zero so $\Delta \text{DE}_\text{head} \approx - DE_{\text{head}}$ when resample ablating most attention heads. As such, for convenience we often plot the original direct effect rather than change in direct effect. For example, consider an attention head with a direct effect of $0.5$ on the correct output token: if this head is ablated and the logit difference between the original and ablated model is only $-0.2$ logits, we can approximate the self-repair as $-0.2 - (-0.5) = 0.3$.

Note that our definition allows for self-repair to be negative. If the original direct effect of an ablated attention head was negative, we expect the self-repair to also be negative. However, if the original direct effect is positive and the self-repair is negative, we believe this should not be thought of as self-repair, but rather downstream breakage where later components perform worse as a result of the ablation. In this work we do not focus on studying downstream breakage.

\subsection{Self-Repair Exists, but Imperfectly}
\label{sec:imperfect-self-repair}
We measure the self-repair of individual attention heads in Pythia-1B \citep{pmlr-v202-biderman23a} on 1 million tokens of The Pile \citep{pile}, the dataset used to train Pythia-1B \footnote{For this experiment and others, unless stated otherwise, we use 1 Million tokens for experiments with the Pythia suite of models and GPT2-Small and GPT2-Medium, but smaller amounts for the other models.}. For a given token and forward pass, we measure each head's direct effect $DE_{head}$ and the outputted logit of the correct next token. We then individually resample ablate each attention head, measure the new logit of the same correct next token, and calculate the logit difference $\Delta \text{logit}$. Averaging these values across the whole dataset, we plot each head's direct effect and logit difference in Figure \ref{fig:self-repair-scatter-main} (see Appendix \ref{across-models-graphs} for the equivalent graphs of other models).

\begin{figure}
%%\vskip 0.2in
\begin{center}
\centerline{\includegraphics[width=0.8\columnwidth]{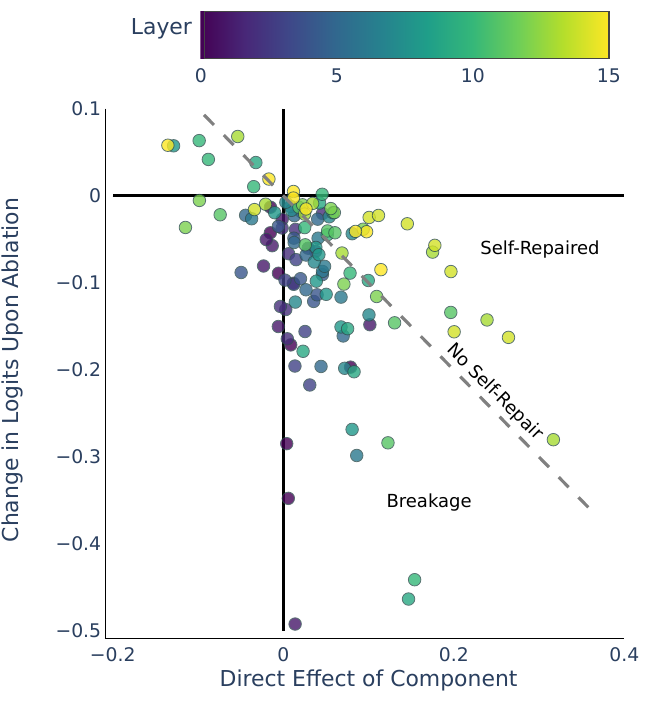}}
%\vskip -0.2in
\caption{Self-Repair of individual Pythia-1B attention heads across 1M tokens on The Pile. For each head in Pythia-1B, we plot its direct effect and the change in logits when resample ablating it.  The heads between the included $y=-x$ line and the x-axis are self-repaired.}
\label{fig:self-repair-scatter-main}
\end{center}
\vskip -0.5in
\end{figure}

Recall that since $\Delta \text{DE}_\text{head} \approx -\text{DE}_\text{head}$, heads which aren't self-repaired have $\Delta \text{logit} \approx -\text{DE}_\text{head}$, as the change directly feeds into the final logits without any downstream compensation. Heads that \textit{are} self-repaired have smaller logit differences, such that $-1 < \frac{\Delta \text{logit}}{\text{DE}_\text{head}} < 0$. In Figure \ref{fig:self-repair-scatter-main}, this corresponds to the heads which fall between the $y=-x$ line and the x-axis. It's clear that there are many such heads, even for heads in the last layer 
(which don't have attention or MLP layers downstream of it\footnote{Pythia-1B is a parallel attention model.}). 

\textbf{The existence of self-repair across the pretraining distribution holds robustly across model sizes and families.} When resample ablating different attention heads in various models, we measure their logit difference $\Delta logit$, change in direct effects $\Delta \text{DE}_\text{head}$ , and thus the self-repair experienced using Equation \eqref{eq:self-repair-eq}.

It has been postulated that each language model component tends to be useful on a sparse set of tokens \citep{bricken2023monosemanticity}, and we corroborate this here by observing a sparsity of significant direct effect. Accordingly, for each head, we measure the average self-repair each attention head experiences across the top 2\% of tokens filtered by each head's direct effect, to ensure we are seeing self-repair of the head's role in the model, rather than noise.

We plot the average self-repair experienced by each attention head in Figure  \ref{fig:self-repair-significance}, along with the mean for all the heads in each layer of a model. It is clear that many heads in later layers of models experience self-repair.

\textbf{Self-Repair is Imperfect:} It's important to note that the heads that \textit{are} self-repaired are not perfectly repaired across the entire distribution. Instead, ablating these heads leads to a small, but noticeable, logit difference. This has important implications for self-repair (see Section \ref{implications}).

%Additionally, the amount of self-repair varies. Self-repair on the direct effect is only meaningful for heads with significant direct effect. Empirically, many early heads don't have strong direct effects, while middle and later heads do.

%This self-repair can often be a significant fraction of the direct effect: if we bound the percentage of direct effect self-repaired on each token between - and 1 to describe the attention head being not self-repaired and losslessly self-repaired, respectively, self-repair can explain up to 89\% of the original direct effect of a layer!
\subsection{Self-Repair is Noisy}
\label{sec:self-repair-noisy}

Self-Repair is a noisy phenomena on multiple levels. Clean hypotheses such as "the direct effects of all late layer heads are self-repaired by 70\%" are immediately falsified (from Figure \ref{fig:self-repair-significance} and \ref{fig:self-repair-scatter-main}). We observed several other phenomena once we moved beyond just averaging over many prompts. These suggest that self-repair is a noisy, difficult-to-study phenomena that is unlikely to have a single clear mechanistic explanation or follow simple rules. These include:

\textbf{Self-Repair varies at the level of tokens:} it's not the case that the self-repair systematically exists similarly across prompts/predictions. Averaging self-repair across the full distribution hides a lot of important detail. Indeed, it is instead extremely noisy.  

\begin{figure}[ht]
%%\vskip 0.2in
\begin{center}
\centerline{\includegraphics[width=0.8\columnwidth]{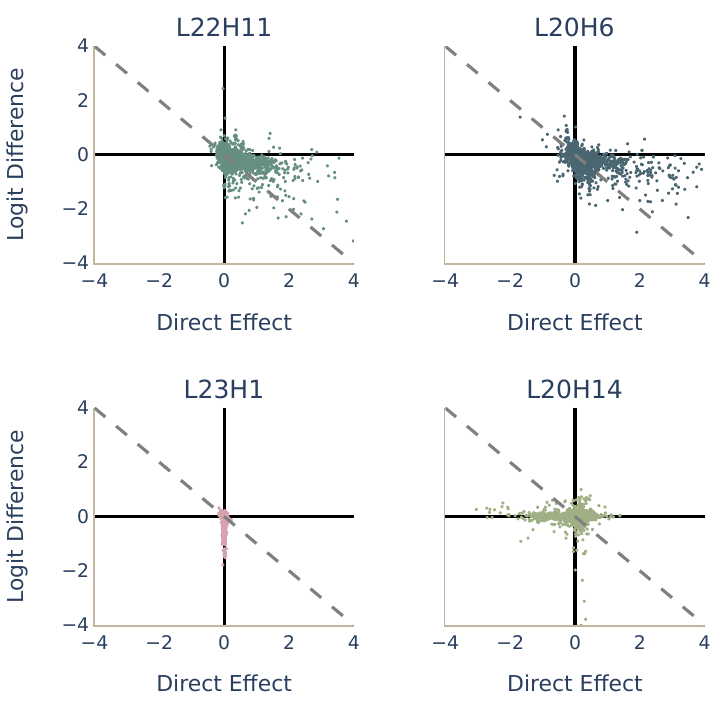}}
\caption{We've handpicked four heads in Pythia-410M, and plotted the direct effect and logit difference when ablating the head across 5000 individual tokens in The Pile. Within a single head, these values can vary highly. The tokens between the included $y=-x$ line and the x-axis are self-repaired.}
\label{single-head-noise}
\end{center}
\vskip -0.3in
\end{figure}

For example, in Figure \ref{single-head-noise} we've plotted the direct effects and logit differences when resample ablating L22H11 (the 11th head in the 22nd Attention Layer) and L20H6 in Pythia-410M on individual tokens. It's clear that there are immense amounts of noise in the self-repair between individual tokens across the same head, despite being able to observe that these two heads appear to be self-repaired on average.

\textbf{Often, the amount of self-repair is far less, or more, than the original direct effect of the head.} If the magnitude of self-repair is quite low, it doesn't quite make sense to declare that self-repair has meaningfully occurred. Additionally, sometimes the measured `self-repair' can be far larger than the original direct effect of the head, which means model performance \textit{improves} upon the ablation of the head. 

\textbf{Many heads in the model don’t have clear correlations between the direct effect of the head and the change in logits upon ablation.} See L23H1 and L20H14 in Figure \ref{single-head-noise} for an example of this. For these heads, there is no clear correlation observed between the direct effect of the head and logit difference when resample ablating it. This lack of a clear correlation was first introduced in \citealt{mcgrath2023hydra} to be a phenomena when ablating layers. We hypothesized that ablating entire layers may have been too drastic of a change (and that smaller-scale ablations may be cleaner): however, even at the level of individual attention heads, this result holds. This is unsurprising for heads without direct effects, but surprising for heads that have significant direct effects yet no strong correlation, such as L20H14: it suggests that the direct effect may be nearly fully self-repaired in some heads.

%Even within the same layer, there exists both heads with and without clear correlations between the direct effects and logit differences.

%\textbf{A given head sparsely contributes to the direct effect.} Normally, a head doesn't contribute, but sometimes it significantly contributes. Thus, self-repair is also sparse in a similar manner: the absolute amount of self-repair may be quite high-variance. Additionally, this means that measuring self-repair as a fraction of the head's direct effect is extremely unstable when the direct effect of a head is low (the fractional amount is higher variance due to noise when dividing by small numbers).

\section{Nontrivial Self-Repair Due to LayerNorm}
\label{section_layernorm}
Given these observations of robust self-repair on the pretraining distribution, what might be some mechanisms behind self-repair? We originally expected for nearly all of the self-repair to occur because of a model's attention heads and MLP layers. However, it turns out that even `passive' components of the model, such as normalizing factors, can induce self-repair.  If we ablate a head and freeze the output of all downstream components \textit{except} for the LayerNorm normalization factor, this recovers a significant logit difference relative to freezing all components \textit{including} the normalization factor.

This is particularly notable because LayerNorm is often treated as a technicality that can be approximated as linear (or ignored) in mechanistic interpretability work \citep{elhage2021mathematical}. We first highlight why to expect self-repair due to LayerNorm, and then highlight empirical results measuring LayerNorm self-repair\footnote{Note that this repair mechanism applies in the same way to RMSNorm, a popular LayerNorm alternative, which is used in LLaMA}.

\subsection{An Argument for LayerNorm Self-Repair}

To provide some intuition as to why LayerNorm could self-repair, consider a simplified case where we resample ablate a head in the last layer of a parallel attention model, such that this change leads only to the final residual stream. 

Naively, we may expect that the change in logits is just $\Delta DE_{\text{Head}}$, as there are no downstream attention heads or MLP Layers. This isn't accurate: LayerNorm acts as a normalizing factor.

As argued in \citealt{elhage2021mathematical}{}, we can simplify the LayerNorm operation such that we `fold in' LayerNorm projections to the weights of linear layers before and after the projection, leaving the nonlinear operation of scaling the residual stream by dividing by $S$, a normalization factor proportional to the residual stream's norm \footnote{Precisely, the normalization factor is calculated by measuring the standard deviation of the residual stream across the residual stream dimension (not the sequence dimension). This is proportional to the norm.}. These scaling factors are a non-trivial change to the forward passes of the model. We can write the logit difference from a resample ablation as the following (which we derive in Appendix \ref{derive-layernorm}):
\vskip -0.2in
\[
\Delta \text{logit} = \underbrace{(\frac{S}{S'} - 1)\text{logit}}_{\text{LayerNorm on \textit{existing logits}}}  + \underbrace{\left(\frac{S}{S'}\right)(\Delta DE_{\text{Head}})}_{\text{LayerNorm on } \Delta DE_{\text{Head}}}
\]
\vskip -0.2in
where $S$ and $S'$ are the scaling factors of the LayerNorm functions  pre- and post- ablation, and $\text{logit}$ is the original logit of the correct next token \footnote{The direct effects in this equation, and throughout our experiments, are measured with respect to the original LayerNorm scaling factors.}. 

Self-Repair occurs because $S > S'$ (as we show empirically in Section \ref{empirical-layernorm}): this means that LayerNorm contributes to self-repair by \textit{amplifying the existing logits}. The LayerNorm scaling also scales $\Delta DE_{\text{Head}}$, but this change is smaller than LayerNorm's impact on the existing logits (as the direct effects are usually smaller in magnitude than the full correct logit).

A possible intuitive explanation for why  $S > S'$ is because each scaling factor corresponds to the norm of the residual stream. This means that ablations decrease the norm of the residual stream. Model components are often correlated: if the output of a head aligns with the residual stream, subtracting it significantly decreases the norm of the residual stream. Since the ablated output of the head is unlikely to align with the residual stream, it won't re-increase the norm. This effect is exaggerated when conditioning on the clean head having a high direct effect, since the head is strongly predicting the correct next token, and likely other model components are too.

\subsection{Empirical Findings of LayerNorm Self-Repair}
\label{empirical-layernorm}
\begin{figure}[ht]
%%\vskip -0.1in
\begin{center}
\centerline{\includegraphics[width=\columnwidth]{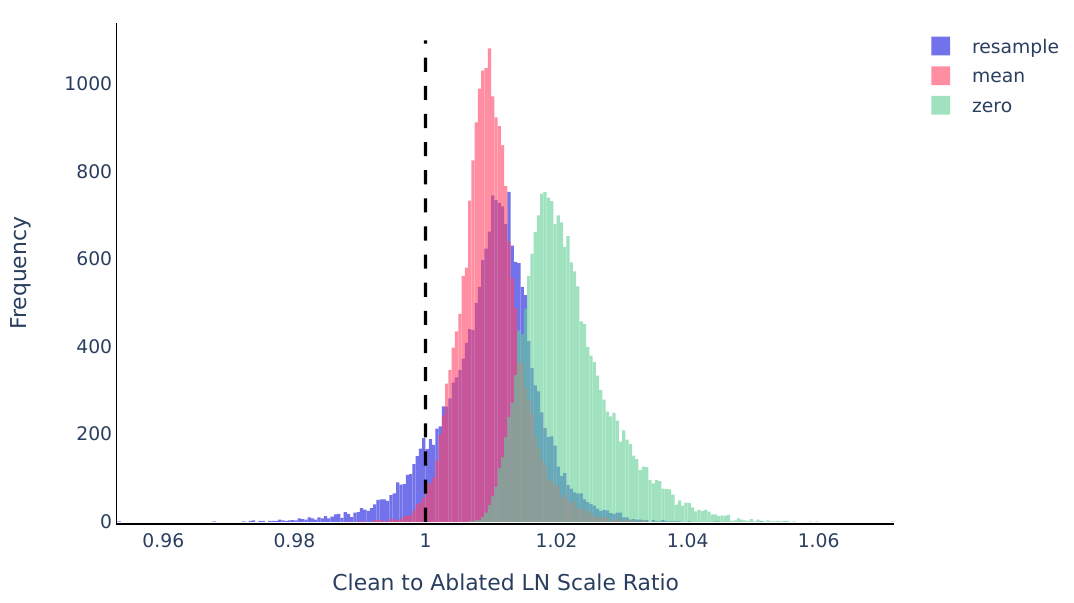}}
\caption{Ratio of clean to ablated LayerNorm scaling factors on L11H2 of Pythia-160M when resample ablating the head over 1 million tokens on The Pile, and then filtering for the top 2\% of tokens according to direct effect. Ratios greater than 1 indicate that LayerNorm is self-repairing by amplifying the existing logits.}
\label{fig:ratio-LN-scales}
\end{center}
\vskip -0.3in
\end{figure}

We empirically confirm that LayerNorm can self-repair via changes in the scaling factor. We take L11H2 in Pythia-160M, which demonstrates this phenomena well, and ablate it across 1 million tokens on The Pile. On the top 2\% of tokens according to its direct effect, we plot the ratio between clean and ablated LayerNorm scaling in Figure \ref{fig:ratio-LN-scales}. We've also included zero and mean ablating, ablations where a head's output is replaced with the zero vector or the average output of the head across a batch, respectively.

On these tokens, the ratio between the clean and ablated LayerNorm scaling factors are almost always greater than one (90.66\% of tokens for resample, 98.75\% for mean, 100\% for zero ablation). This increase causes self-repair (even for a ratio such as 1.02, see Appendix \ref{app:ln-ratios})!

\begin{figure}[ht]
%%\vskip 0.2in
\begin{center}
\centerline{\includegraphics[width=\columnwidth]{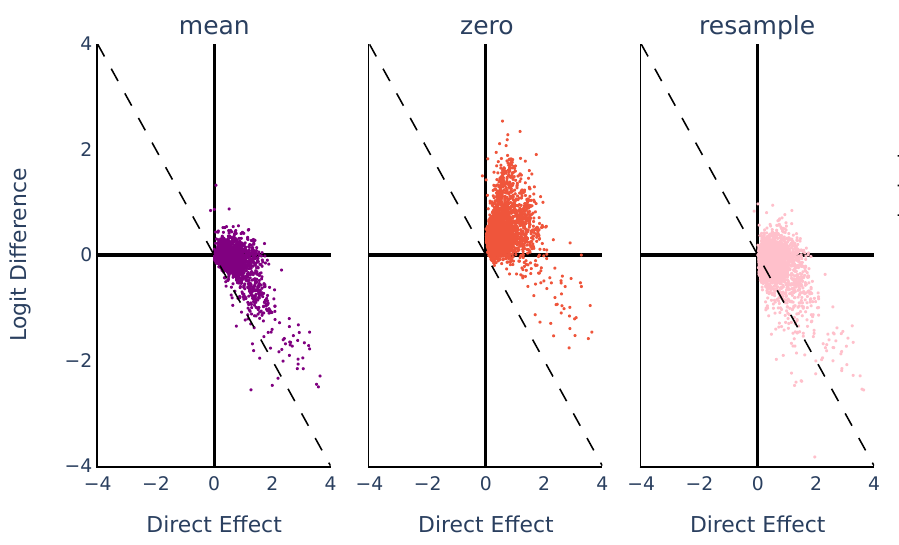}}
\caption{Direct effect vs logit difference of L11H0 in Pythia-160M under different ablations. Notice how zero ablations can induce positive logit differences. Recall that this self-repair can only occur due to LayerNorm scale changes.}
\label{ablation-comparison}
\end{center}
\vskip -0.3in
\end{figure}

This highlights how various forms of common ablations (zero, mean, and resample) can sometimes influence the scaling factors quite significantly. In particular, zero ablating may strongly change the norm of the residual stream. This has practical consequences for the self-repair of attention heads: Figure \ref{ablation-comparison} highlights the self-repair that occurs ablating L11H10 in Pythia-160M (selected as it shows our point well) with the different ablations. Recall that L11H10 is in the last attention layer and that Pythia-160M is a parallel-attention model: as such, the only component responsible for any self-repair is LayerNorm. Zero ablating can \textit{increase the logit difference}, even when ablating the head while it has a positive direct effect. Though, zero ablating doesn't always have an extreme effect, see Appendix \ref{appendix:ablations-change-norms}.

%In practice, LayerNorm self-repair is usually small, but not trivial. Again, the type of ablation you perform can also influence how much LayerNorm self-repair occurs. For instance, zero-ablations more strongly reduce the norm of the model, as shown below %TODO: figure here

To attempt to quantitatively measure the extent to which LayerNorm self-repairs, we break down the self-repair experienced by each head into LayerNorm, MLP, and attention head components as follows:
\begin{equation}
\label{eq:self_repair_breakdown} \\
\begin{aligned}[t]
\text{self repair} &= \Delta \text{logit} - \Delta \text{DE}_{\text{head}}  \\
                   &= \sum_{h\in H}\Delta \text{DE}_{h} + \sum_{m\in M}\Delta \text{DE}_m + \Delta \text{DE}_{\text{LayerNorm}}
\end{aligned}
\end{equation}

where we are measuring the sum of the:
\begin{compactenum}
    \item Changes in direct effect $\Delta\text{DE}_{h}$ for each head $h$ in the set $H$ of all heads downstream of the ablated head.
    \item Changes in the direct effect $\Delta\text{DE}_m$ for each MLP layer $m$ in the set $M$ of all MLP layers downstream of the ablated head.
\end{compactenum}
and using the difference between the calculated self-repair and this sum to determine how much of the remaining self-repair is due to changes in the final LayerNorm scale $\Delta \text{DE}_{\text{LayerNorm}}$. This ends up accounting for the effect of the LayerNorm scale on both the existing logits and the changes in all direct effects.

\begin{figure}[ht]
%\vskip -0.2in
\begin{center}
\centerline{\includegraphics[width=\columnwidth]{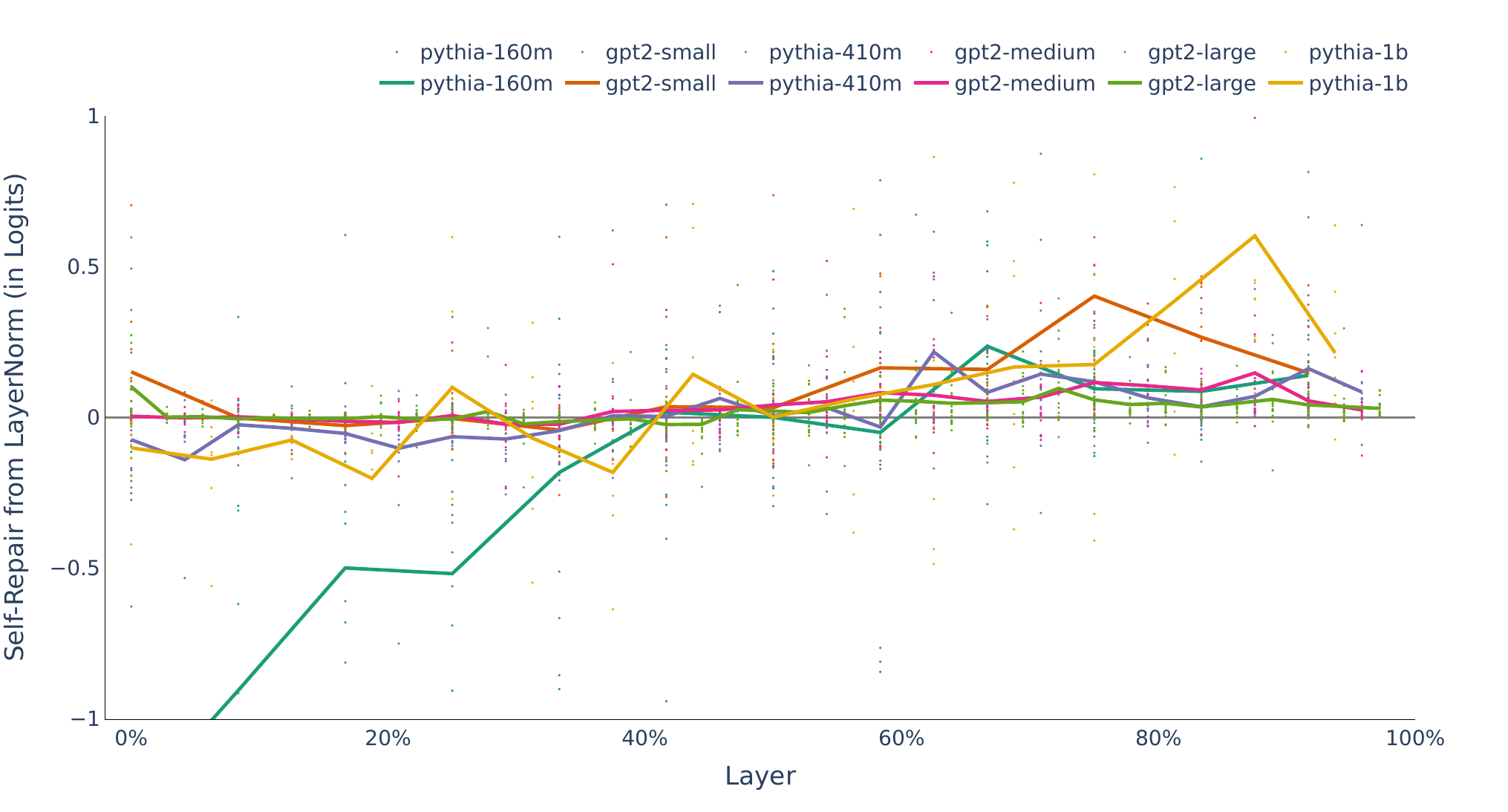}}
\caption{Across the top 2\% of tokens on The Pile, we measure the LayerNorm self-repair when resample ablating individual heads. We've plotted mean lines averaging the values across each model's layer. LayerNorm has a significant effect on later layers of the model.}
\label{fig:general-ln-self-repair}
\end{center}
\vskip -0.4in
\end{figure}

Across models, we calculate these values from Equation \ref{eq:self_repair_breakdown} for each attention head on the top 2\% of tokens in The Pile according to their direct effect. The self-repair due to LayerNorm is plotted in Figure \ref{fig:general-ln-self-repair}.

As self-repair is noisy (Section \ref{sec:self-repair-noisy}), creating a summary statistic to capture how much self-repair explains the direct effect of a head is extremely difficult. Measuring self-repair as a fraction of the direct effect often creates extreme values that are uninterpretable and not useful.

But, when measuring the self-repair due to LayerNorm on a token as a fraction of the clean direct effect of the head on that token, \textit{and capping the percentage between 0 and 100\%}, we learn that LayerNorm can explain around \LN of the direct effect of a head on average across many layers. This is best treated more as a directional metric highlighting the existence of this motif, rather than a concrete, precise one; we illuminate the difficulty in precisely measuring self-repair as fractions of direct effects, and our approaches towards dealing with this, in more detail in Appendix \ref{appendix:summary-stat}.

\section{Sparse Neuron Anti-Erasure Helps Self-Repair}
\label{Section_AntiErasure}
\citealt{mcgrath2023hydra} identified MLP Erasure, a behavior in MLP layers where later MLP components partially negate important directions outputted by earlier components. We study this in the context of direct effects: the direct effect of components will occasionally be negated by downstream MLP layers. Importantly, Erasure was defined when there is a causal link between the early components and the downstream components that partially negate them\footnote{We can imagine a case where there is a third component upstream of both the positive and negative component that causes both effects, with no causal link between the positive and negative component. This would not be erasure.}. 

As an example of Erasure, imagine an early component $P$ which contributes $x$ to a certain logit, and a later component $N$ casually downstream of $P$ which contributes $-0.7x$ to that same logit, resulting in a net effect of $0.3x$. The direct effect of $P$ is $x$ (as this ignores $N$), but if we zero ablate $P$'s output, then this may also change $N$'s net effect to zero, and thus the total change in logits is just $0.3x$ - there is $0.7x$ logits of self-repair!

We call this self-repair "Anti-Erasure",  where the removal of the upstream component's direct effect also induces the removal of a causally downstream Erasure. A similar motif was discovered in Copy Suppression attention heads \citep{mcdougall2023copy}, where the ablation of specific attention heads caused the Copy Suppression heads to perform less Erasure. We build upon this initial work to demonstrate that Anti-Erasure is an important mechanism in sets of sparse neurons of MLP layers that self-repairs.

\subsection{Erasure Occurs in Neurons}

\citealt{mcgrath2023hydra} introduced MLP Erasure as a behavior performed by MLP layers that caused self-repair. In order to better understand this phenomena, we narrowed our focus from the broad unit of analysis of MLP layers to the level of neurons. We find that specific neurons perform Anti-Erasure and thus help self-repair.

\begin{figure}[ht]
%%\vskip 0.2in
\begin{center}
\centerline{\includegraphics[width=\columnwidth]{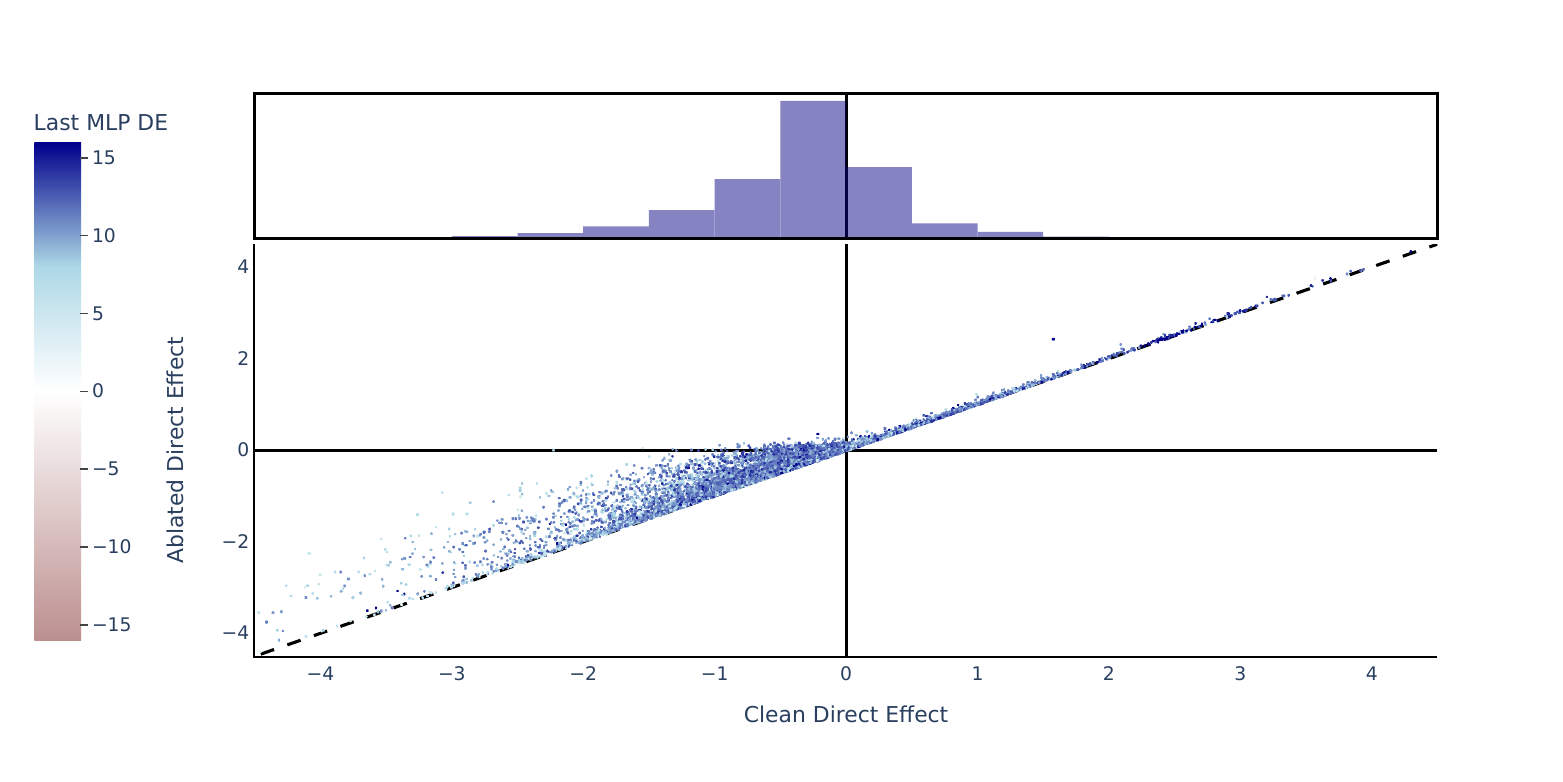}}
\caption{On select tokens, we plot the clean and ablated direct effect of the top self-repairing neuron in the last MLP layer when resample ablating L10H11 in Pythia-160M (and include a histogram of the clean direct effects of the neurons). Each token is colored with the direct effect of the entire MLP layer on the clean run. The majority of the neurons are performing Anti-Erasure. Additionally, despite the MLP layer almost always contributing positively to the direct effect, the top self-repairing neuron typically performs Anti-Erasure and has a negative clean direct effect.} %
\label{Neurons_anti_erasure}
\end{center}
\vskip -0.3in
\end{figure}

A head that highlights this well is L10H11 in Pythia-160M, an attention head right before the final MLP layer. We run the model on a subset of The Pile, resample ablate the head, and measure the self-repair on each token. Then, we filter for the top 2\% of instances in which the last MLP layer self-repairs the most. As the output of the MLP layer is the sum of the output of each neuron, we can also measure the self-repair of each neuron: as such, for each token, we isolate the individual neuron that self-repairs the most in the layer (for that instance) and plot its clean and ablated direct effect, colored by the direct effect of the entire last MLP layer, in Figure \ref{Neurons_anti_erasure}. 

We observe that across many tokens, the top self-repairing neuron has a negative clean direct effect and a less negative ablated direct effect upon ablation of L10H11. This indicates that it was originally performing erasure, but performed less erasure upon the ablation of L10H11. This is in contrast to the direct effects of the entire last layer, which are frequently extremely positive (as indicated by the darker blue color of the majority of the points in Figure \ref{Neurons_anti_erasure}).

\subsection{Sparse Sets of Neurons Perform Anti-Erasure}

\begin{figure}[t]
%\vskip -0.2in
\begin{center}
\centerline{\includegraphics[width=\columnwidth]{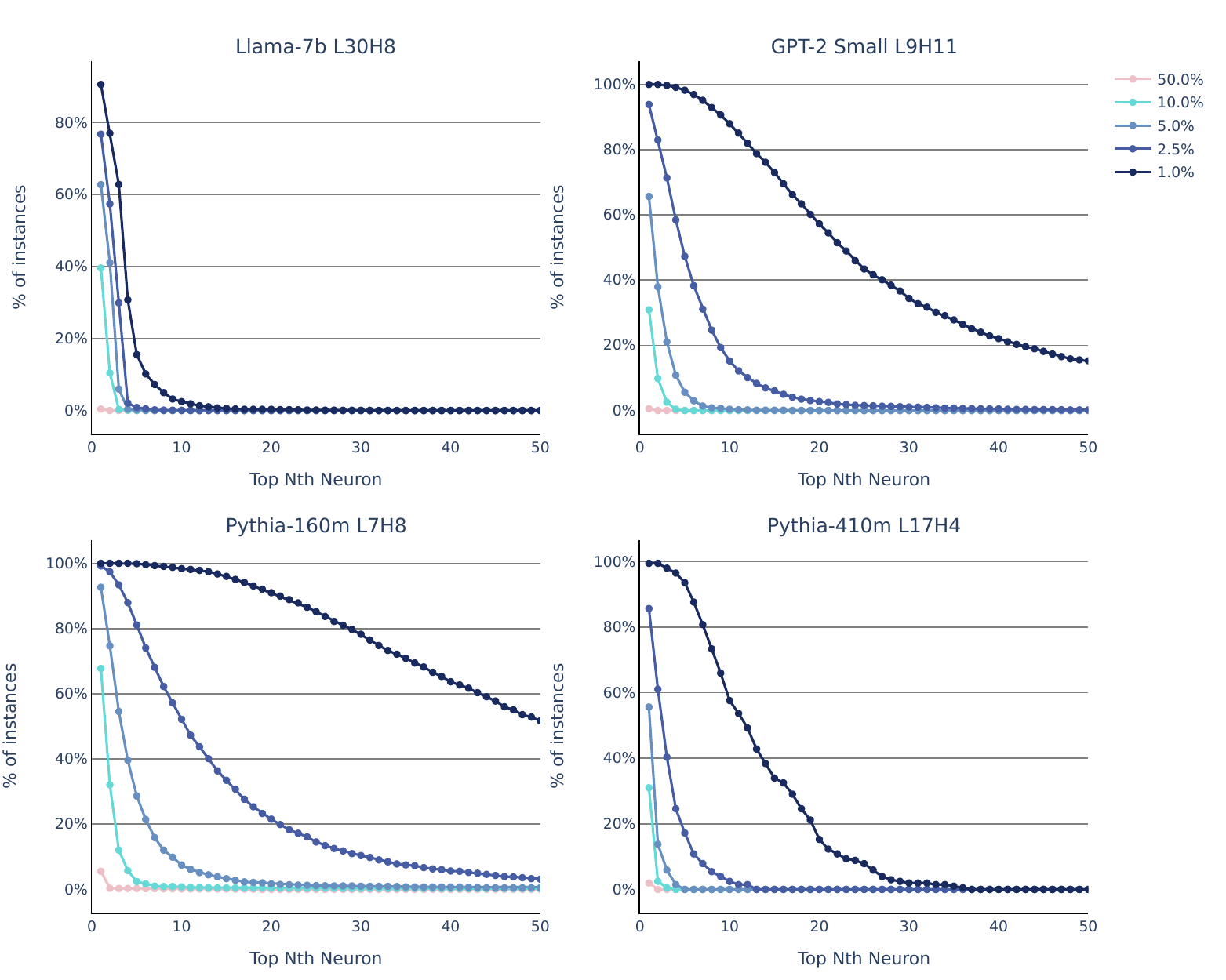}}
\caption{For four models and random attention heads, we plot the frequency at which the top $N$th self-repairing neuron repaired X\% of the direct effect (on the top 2\% of the tokens by direct effect).}
\label{three_neurons_needed}
\end{center}
\vskip -0.4in
\end{figure}

In most instances of self-repair, a sparse number of neurons have significant amounts of self-repair. For four different models, we resample ablate a specific head in a later layer, and see how much the top self-repairing neurons self-repair. We select heads before the final MLP layer which have a large average direct effect across tokens from The Pile\footnote{To ensure this was a representative sample, once we selected the heads, we stuck with these heads throughout our analysis.}. For each head, we isolate the 2\% of tokens on The Pile for which the head has the highest direct effect and measure the self-repair that occurs due to each of the neurons in the final MLP layer. We can further divide by the head's direct effect to get the fraction of the direct effect compensated for by each neuron. To gauge how much self-repair is explained by a sparse subset, we sort and consider each of the top 50 self-repairing neurons, and plot the percentage of instances for which the \textit{neuron} explained 50, 10, 5, 2, and 1\% of the direct effect of the head in Figure \ref{three_neurons_needed}. For instance, in Llama-7B, the top self-repairing neuron repairs 10\% of the direct effect of L30H8 in over 40\% of instances.

It's clear that a sparse number of neurons self-repair heavily: recall that Llama-7B \citep{touvron2023llama}, GPT2-Small \citep{Radford2019LanguageMA}, Pythia-160M, and Pythia-410M have 11008, 3072, 3072, and 4096 neurons per layer, respectively. Frequently, these are Erasure neurons (shown further in Appendix \ref{appendix:changes-mlp-neurons}). As such, frequently, a sparse set of anti-erasure neurons have significant amounts of self-repair\footnote{We considered developing a summary statistic to capture this effect, but realized the eventual result could quickly change from slight changes in the experimental setup. However, anecdotally, it was frequently the case that self-repair of a few of the top self-repairing neurons could sum to equal the vast majority of the entire layer's self-repair.}.

\subsubsection {How Important is the Sparse Anti-Erasure?}
Although a sparse number of neurons self-repair significantly, this doesn't imply that these are the only self-repairing neurons. On a given ablation, many neurons have marginally decreased or increased direct effects. This means that while a small subset of the top self-repairing neurons may sum up to the `total self-repair',  many other neurons may have changed direct effects that cancel each other out. In particular, when sorting, the sum of the largest fractions of self-repair may significantly exceed $1$ if they are compensated for by many neurons with reduced direct effects. 

We suspect that when ablating the same attention heads, roughly 60\% of the neurons in the final layer have negligible changes in direct effect and that 40\% of the neurons have changes in self-repair which are small and cancel out, yet are non-negligible when considering their combined absolute change in direct effects (Appendix \ref{appendix:changes-mlp-neurons}). This indicates that the full picture of self-repair in the entire MLP layer is more complex and that sparse Anti-Erasure is only one part of a larger story: it's not the case that all the neurons stay stagnant besides a set of anti-erasure neurons. Instead, while sparse sets of anti-erasure neurons may help fuel large amounts of self-repair, there is also a significant amount of neurons that have smaller changes in their direct effect.

%The self-repair of the top neurons may be significant on a specific instance, but this doesn't mean that they are the \textit{only} self-repairing neurons: there may be many other neurons with changed direct effects, both positively and negatively.

\subsubsection{Anti-Erasing Neurons Differ Across Prompts}
Are there neurons that perform Erasure and self-repair consistently? Anecdotally we observed that the same neurons can often self-repair across the same prompt, but across different prompts, they may differ. We filter for the top 2\% of tokens where L10H11 in Pythia-160M is self-repaired the most and collect the top 10 repairing neurons in the last layer, per prompt. The most any single neuron is in the top-10 self-repairing neurons is 16\% of the tokens.

This suggests that there are likely different neurons responsible for different forms of erasure. In particular, this is potentially related to "Suppression Neurons", discovered to decrease probabilities of related tokens \cite{gurnee2024universal}. This would explain why neurons may similarly self-repair across the same prompt, but not on the entire distribution.

%However, it's unclear if the presence of specific tokens are the only thing that are erased. In preliminary experiments which isolated the outputs of attention heads to output only in the direction of the next token, or only output everything else, self-repair occurred in both cases. As such, this suggests that the anti-erasure effect isn't just a naive "erase token" behavior. 
% \begin{figure}[ht]
% %%\vskip 0.2in
% \begin{center}
% \centerline{\includegraphics[width=\columnwidth]{figures/different_neurons_self_repair.png}}
% \caption{Different neurons are performing self-repair}
% \label{different_neurons}
% \end{center}
% %%\vskip -0.2in
% \end{figure}
\section{Discussion}
In this section, we discuss some of the consequences of self-repair. We also outline the Iterative Inference Hypothesis, which we believe may help describe portions of self-repair.

%In this section, we aim to present a series of motifs which we hope provides a high-level picture into why self-repair may manifest, and highlight some practical consequences of them. None of these individually provide a sufficient explanation for all of self-repair, but we expect that all of these combined are motifs which will ultimately explain a significant fraction of self-repair.
\subsection{Implications of Imperfect Self-Repair for Interpretability Efforts}
\label{implications}

The major problem with self-repair for interpretability efforts is that it makes ablations an unreliable tool: we would naively hope that if we ablate a model component and look at the effect on model performance, this change would accurately capture the component's role in the model. However, if the component experiences significant self-repair, the direct effect and change in logits that occur upon ablation of the component may be very different. In fact, it becomes unclear what the "true effect" of the component even is - if a component is perfectly self-repaired yet has a high direct effect, does this mean it is irrelevant or important yet compensated for?

In practice, this is most concerning when considering circuit analysis, where we patch or ablate individual model components and observe their effect on the output, in order to isolate the sparse subgraph of the model relevant to our task, as performed in \citep{wang2023interpretability,lieberum2023does,conmy2023towards}. However, fortunately, circuit analysis only requires identifying whether a component is important or unimportant (i.e. whether it belongs in the sparse subgraph), not the precise effect of ablating it. So the fact that self-repair is imperfect across the general distribution (Section \ref{sec:imperfect-self-repair}) helps reduce some of the concerns for circuit discovery efforts; because the `importance' of various components is extremely heavy-tailed, even a significant fractional decrease in the estimated effect won't change which nodes are important.

However, this doesn't fully alleviate all concerns. In certain situations, self-repair can be lossless or overcompensate. This may happen on certain narrow distributions or may be induced depending on what tools you use. And if the degree of self-repair differs significantly between components, borderline components may be incorrectly included or excluded.

% Although self-repair exists, the fact that it is imperfect across the general distribution (Section \ref{sec:imperfect-self-repair}) helps reduce some of the concerns for automated circuit discovery efforts. C aims to find a meaningful sparse subgraph for a given task, i.e. finding important components  \citep{conmy2023towards}. This is normally sufficiently heavy-tailed such that even a significant fractional decrease in the estimated effect won't change which nodes are important.

% However, this doesn't fully alleviate all concerns. In certain situations self-repair can be lossless or overcompensate. This may happen on certain narrow distributions, or may be induced depending on what tools you use.

\subsection{Taking Models Off-Distribution can Cause Unexpected, Easily Misinterpretable, Consequences}

One reason to expect self-repair to occur is that taking the model off-distribution can lead to very surprising consequences. Ablations fundamentally take the model off-distribution, in the sense that the set of internal activations achieved from ablations may be impossible to achieve on \textit{any} input. Models were not trained to respond coherently to these kinds of internal interventions, and so may behave in erratic and hard-to-predict ways which are difficult to reason about.

When we first found self-repair, we assumed that models were intentionally self-repairing. However, the LayerNorm-based self-repair seems to be a side effect of unrelated mathematical properties of the model, interacting with how our ablations change the norm of the residual stream. It's possible that many observed instances of `self-repair' are just uninteresting consequences of throwing the model off-distribution, for which this will be clear after discovering a few more insights like the above.

Importantly, the fact that self-repair occurs across models tells us something about the internal mechanisms of these models, but it may be a byproduct of some other mechanism in the model, rather than that they are intentionally self-repairing. 

%For instance, LayerNorm explaining portions of self-repair isn't an representation of self-repair that the model has learned, but rather a random byproduct of throwing the model norm off-distribution. 
 %Even on 'interpretable' instances, ablations can often lead to extremely strange, uninterpretable, consequences: for instance, we once discovered that Pythia-160M consistently acts very strangely on resting positions, and that ablating on these positions yielded strange results. 

This is a problem fundamental to ablations and is likely difficult to circumvent. A suggested course of action is to try to control how far from the standard distribution one's causal interventions take the model \cite{chan2022causal}. Two ways this can be done is by avoiding zero ablation (which most significantly adjusts the residual stream norm, as shown in Section \ref{empirical-layernorm}) or by freezing LayerNorm while ablating. Additionally, one could use path patching \citep{goldowskydill2023localizing} instead of patching full components.

Our results do not demonstrate that all of self-repair is a byproduct of other mechanisms: LayerNorm self-repair only explains up to around \LN of the self-repair on average. But these results are a word of caution to be careful when taking models off distribution.

% \subsection{Erasure is Useful, but Self-Repairing}
% From our discussion of MLP Erasure in Section \ref{Section_AntiErasure} to the introduction of Copy Suppression in \citealp{mcdougall2023copy} it appears to be the case that models have erasure components throughout them. Regardless of why Erasure exists, it appears to be a useful property models utilize, and thus indicates we can expect for self-repair of the form of Anti-Erasure to be a common motif across models.

% We expect that identifying these Erasure components and then freezing them during circuit analysis can help control for downstream self-repair and still largely identify many important components of the model.

\subsection{Iterative Inference Hypothesis}
\label{sec:model-iterativity}
There's a wealth of evidence \citep{logit_lens, dar2023analyzing} that models gradually build up their final logits in unembedding space: the correct token isn't predicted by a singular model component, but rather built up over time by the outputs of multiple components. An implication of this is that models can still be relatively accurate, even when removing the final layer. %Naively, this is surprising: the final layer is  the final layer is always there, so why are models robust to removing it? 

One perspective from which self-repair becomes less surprising is what we call the Iterative Inference hypothesis: rather than layers being part of a top-down process, assisting all future layers in complex circuits, many model components are more of a bottom-up process, where each layer treats the input as a guess for the final output, and tries to reduce the error between this guess and the true next token. 

From this perspective, self-repair is unsurprising. Imagine that some task, such as Name Moving, must be performed. Some earlier component in the model will write the signal into the residual stream that a task $T$ needs to be completed. If the Iterative Inference Hypothesis holds, then rather than there being some dedicated head to do the task, there are many such capable heads: the earliest one that reads in the signal will perform the task $T$, and likely write the signal that the task is complete. Importantly, if the head that performed task $T$ was ablated, then the need to complete task $T$ would still exist (as well as the associated information of the signal to do so): as such, it is possible for another downstream head to observe and complete it instead.

%We highlight evidence for the Iterative Inference Hypothesis in Appendix \ref{additional-discussion-IIH}, including Self-Repressing heads and the existence of similar heads across layers. 
\subsubsection{Evidence for, and Discussion of the Iterative Inference Hypothesis}
\label{additional-discussion-IIH}
What supports the Iterative Inference Hypothesis? Firstly, the hypothesis is consistent with the evidence presented in \citealt{mcdougall2023copy} on how specific heads can influence downstream heads to not perform a task. Another line of evidence that supports this hypothesis is the identification of attention heads across models which we dub `self-reinforcing' and `self-repressing'.  

For a specific attention head, we take its output on a forward pass and re-run the forward pass while adding the output of the head \textit{back into the residual stream which feeds into the head}. We measure the original and new direct effects as a result of this intervention.

\begin{figure}[ht]

\begin{center}
\centerline{\includegraphics[width=0.75\columnwidth]{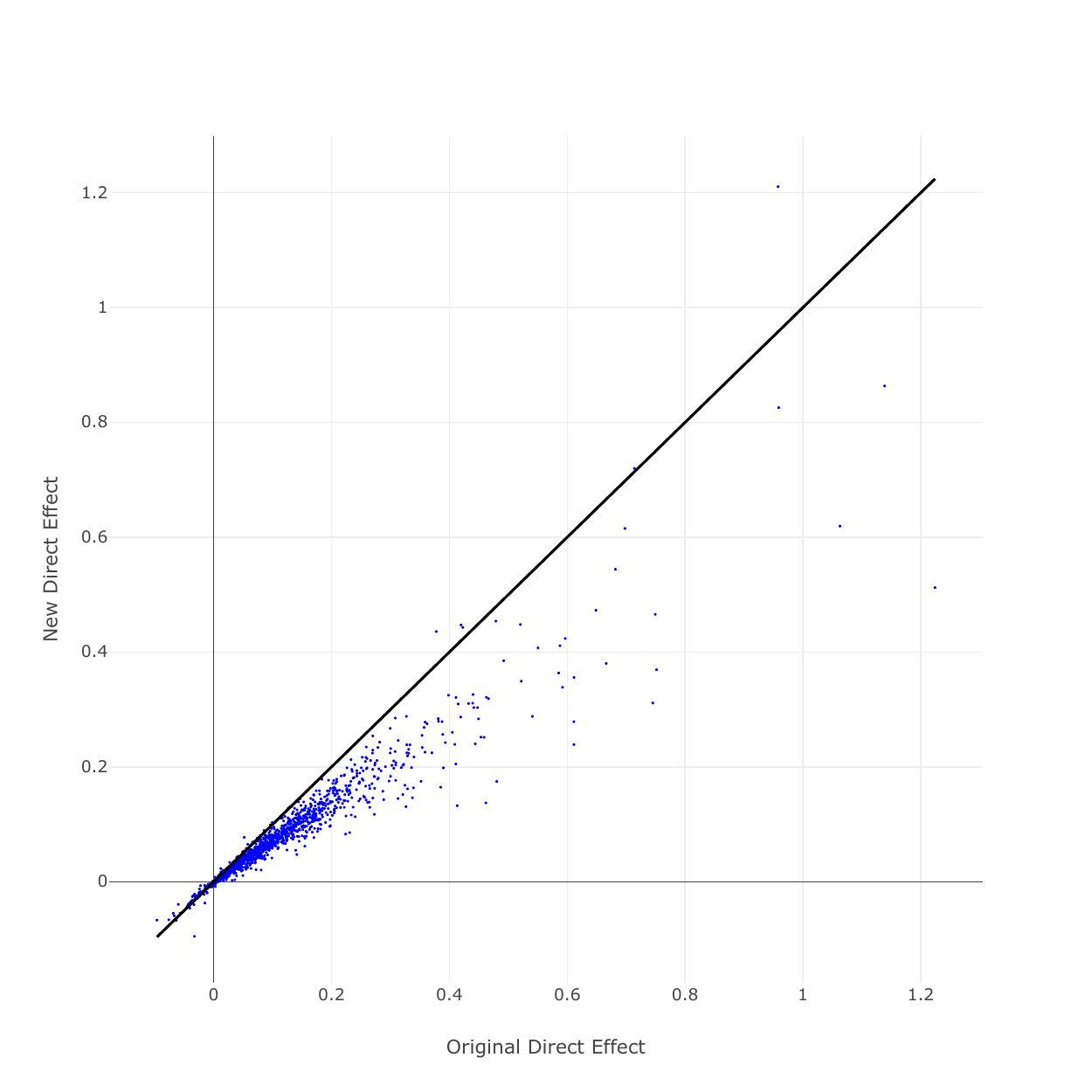}}
%\vskip -0.1in
\caption{A self-repressing head in GPT2-Medium is L21H1. We plot L21H1's original and new direct effects on different tokens after adding its output, scaled by three, back into the previous residual stream. L21H1 has a decreased direct effect as a result.}
\label{self-repressing}
\end{center}
\vskip -0.3in
\end{figure}

We observe `self-reinforcing' and `self-repressing' heads, in which the direct effect of the head increases or decreases proportional to how much of the original output of the head is added. Figure \ref{self-repressing} highlights one head in GPT2-Medium which `self-represses'. The repression is proportional to the amount the output is scaled (Appendix \ref{appendix:self-repress-scaling}).

Not all heads are self-reinforcing and self-repressing. Some appear to be combinations of the two. However, the presence of self-repressing heads suggests that certain attention heads may be outputting a signal that specifies that the task they performed is not needed anymore.

An additional reason to think that the Iterative Inference hypothesis is true is that different attention heads in different layers can perform similar tasks.

The Indirect Object Identification task \citep{wang2023interpretability} highlights one of the earliest instances of observed self-repair. In it, some attention heads are classified as "Name Mover Heads" and other heads as "Backup Name Mover Heads". Originally, it was believed that these were meaningfully different components: however, when running the model across a more general distribution, it is possible to observe instances of the Backup Name Mover Heads performing name moving as well. 

The "Name Mover Heads" exist in earlier layers than the "Backup Name Mover Heads". The fact that these "Backup" heads can perform moving behavior suggests that they have the capability to perform Name Moving, but don't do so in the Indirect Object Identification task. As a result, ablating the Name Mover Heads would additionally preserve the associated signal to perform Name Moving, which the Backup Heads read in and perform as a result.

Additionally, when moving the residual stream read by the "Name Movers" into the residual stream read by the "Backup Name Movers", the backup heads begin performing name moving as well (note that this is equivalent to zero ablating all the attention heads and MLP layers in between them). This further highlights the likelihood of there being a shared stimulus to perform Name Moving which both types of heads respond to, but which the Name Movers often respond to and ablate.

However, there is also a wealth of evidence against this hypothesis. Anecdotally, previous token and induction heads \citep{olsson2022context} don't seem to be self-repaired. Perhaps this may be because these heads are always expected to perform these `fundamental' tasks across all prompts. Further, prior literature has emphasized the presence of complex circuits within the model, which don't easily exist across layers, the S-inhibition heads in the Indirect Object Identification task \citep{wang2023interpretability} being one of them. 

This highlights how specific tasks in the head may not be performed `iteratively': they may be too fundamental such that the model has specific heads dedicated to them, or it may be too complex/costly to perform iteratively by multiple heads. Many heads in the model may not be performing `iterative' tasks.

\section{Related Work}
\label{sec:related-works}

The \textbf{self-repair phenomena} was first identified in the Indirect Object Identification distribution \cite{wang2023interpretability}. Self-Repair was initially explored as the Hydra Effect \cite{mcgrath2023hydra}, and a specific instance of it was explained by Copy Suppression \cite{mcdougall2023copy}. Our work differs from this prior work as follows:

\begin{compactenum}
    \item We analyze self-repair across the a pretraining distribution and with individual attention heads. Past work has mostly highlighted the presence of self-repair in narrow distributions or when ablating larger components.
    \item Our introduction of LayerNorm is as a self-repairing mechanism is an entirely novel contribution.
    \item  Prior work has identified self-repairing Anti-Erasure in entire MLP layers \citep{mcgrath2023hydra} or discussed Anti-Erasure in the context of Suppression heads \citep{mcdougall2023copy}. Our work highlights how erasure in MLP layers is better understood in terms of ‘erasure neurons’, which are sparse and self-repair significantly.
\end{compactenum}

Understanding self-repair fits in the broader work of \textbf{Mechanistic Interpretability}, which aims to reverse engineer neural networks. Related work includes \citealt{olah2020zoom} on vision models and \citealt{meng2023locating} on Transformer models. Previous work has attempted to understand the behavior of individual neurons \citep{Bau_2020, gurnee2023finding} or attention heads \citep{gould2023successor}. In particular, suppression neurons were previously found in \citealt{voita2023neurons} and \citealt{gurnee2024universal}. One important aspect of Mechanistic Interpretability is automating circuit discovery \citep{conmy2023towards, bills2023language}. 

Recent research has emphasized the importance of using casual mechanisms to measure component importance \citep{chan2022causal}. Ablations have been a common technique for this \citep{leavitt2020falsifiable}, and have been used to validate hypothesis in \citealt{olsson2022context, nanda2023progress}.

Ideas related to the \textbf{Iterative Inference Hypothesis} were introduced in \citealt{greff2017highway, jastrzębski2018residual}. The Universal Transformer \cite{dehghani2019universal}, Logit Lens tool \cite{logit_lens}, and Tuned Lens tool \cite{belrose2023eliciting} were built on ideas similar to this.

A related phenomena is \textbf{parameter redundancy}, focused on reducing parameters without harming performance. Self-repair is a different perspective, but may explain how one can ablate individual attention heads without significant performance loss, as demonstrated in \cite{michel2019sixteen}.

\section{Conclusion}
In this work, we've highlighted how attention head self-repair exists across entire pretraining distributions and models. We've explored different mechanisms that self-repair, including LayerNorm self-repair and sparse neuron Anti-Erasure.

A strong limitation of our work is that it focuses on the repairing of direct effects as opposed to other `behavioral' measures of attention heads (which aren't immediately represented as direct effects). It's unclear how much the mechanisms highlighted in sections \ref{section_layernorm} and \ref{Section_AntiErasure} will generalize to these other potentially viable forms of self-repair. Despite this, we expect that the ideas presented here will help provide some grounding for further research on self-repair.

\section*{Impact Statement}
This paper aims to advance the field of Artificial Intelligence interpretability. We hope these insights help build tools to make uninterpretable machine learning systems more interpretable, and thus more understandable, safe, and controllable.

\section*{Acknowledgements}
Thanks to Jonathon Schwartz and Diana Leung for presenting some preliminary results which were used in section \ref{additional-discussion-IIH}. Jonathon Schwartz was particularly helpful in providing helpful comments on the paper. Thanks to Wes Gurnee, Arthur Conmy, Tom McGrath, and Swarat Chaudhuri for useful discussions. Portions of this work were supported by the MATS program as well as the Long Term Future Fund. Additionally, this work was supported by the Center for AI Safety Compute Cluster. Any opinions, findings, mistakes, conclusions or recommendations in this material are our own and do not necessarily reflect the views of our sponsors or employers.

\section*{Author Contributions}
     Cody Rushing directed the project, conducting all experiments and writing the majority of the paper. Neel Nanda supervised this project, providing helpful advice and support at all stages in the project.

%\textbf{Do not} include acknowledgements in the initial version of the paper submitted for blind review.

% If a paper is accepted, the final camera-ready version can (and
% probably should) include acknowledgements. In this case, please
% place such acknowledgements in an unnumbered section at the
% end of the paper. Typically, this will include thanks to reviewers
% who gave useful comments, to colleagues who contributed to the ideas,
% and to funding agencies and corporate sponsors that provided financial


\bibliography{self_repair_paper}
\bibliographystyle{icml2024}

%%%%%%%%%%%%%%%%%%%%%%%%%%%%%%%%%%%%%%%%%%%%%%%%%%%%%%%%%%%%%%%%%%%%%%%%%%%%%%%
%%%%%%%%%%%%%%%%%%%%%%%%%%%%%%%%%%%%%%%%%%%%%%%%%%%%%%%%%%%%%%%%%%%%%%%%%%%%%%%
% APPENDIX
%%%%%%%%%%%%%%%%%%%%%%%%%%%%%%%%%%%%%%%%%%%%%%%%%%%%%%%%%%%%%%%%%%%%%%%%%%%%%%%
%%%%%%%%%%%%%%%%%%%%%%%%%%%%%%%%%%%%%%%%%%%%%%%%%%%%%%%%%%%%%%%%%%%%%%%%%%%%%%%
\newpage
\appendix
\onecolumn

\section{Troubles with Quantifying Self-Repair Relative To Direct Effects}
\label{appendix:summary-stat}

As we highlighted in Section \ref{sec:imperfect-self-repair}, on many occasions the direct effect of a head is sparse. This means that often, measuring self-repair as a fraction of the total direct effect can lead to extreme values which, when averaged with others, skew the data immensely. 

\begin{figure}[h]
\centering

\begin{minipage}{0.45\linewidth}
\centering
\includegraphics[width=1\linewidth]{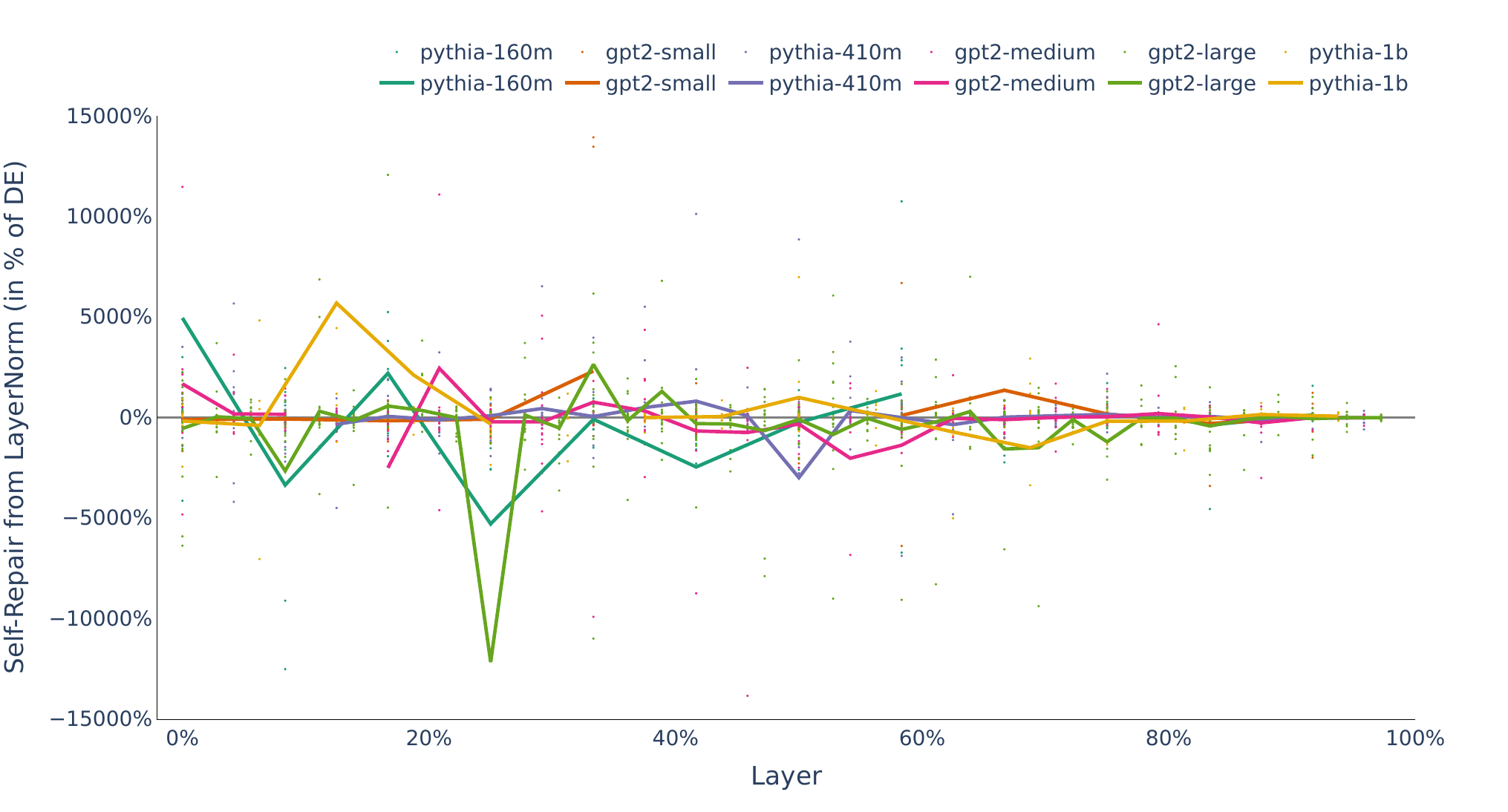}
\caption{'Unclipped' measuring of LayerNorm as a fraction of direct effect}
\label{fig:extreme-values}
\end{minipage}
\hfill
\vline
\hfill
\begin{minipage}{0.45\linewidth}
\centering
\includegraphics[width=1\linewidth]{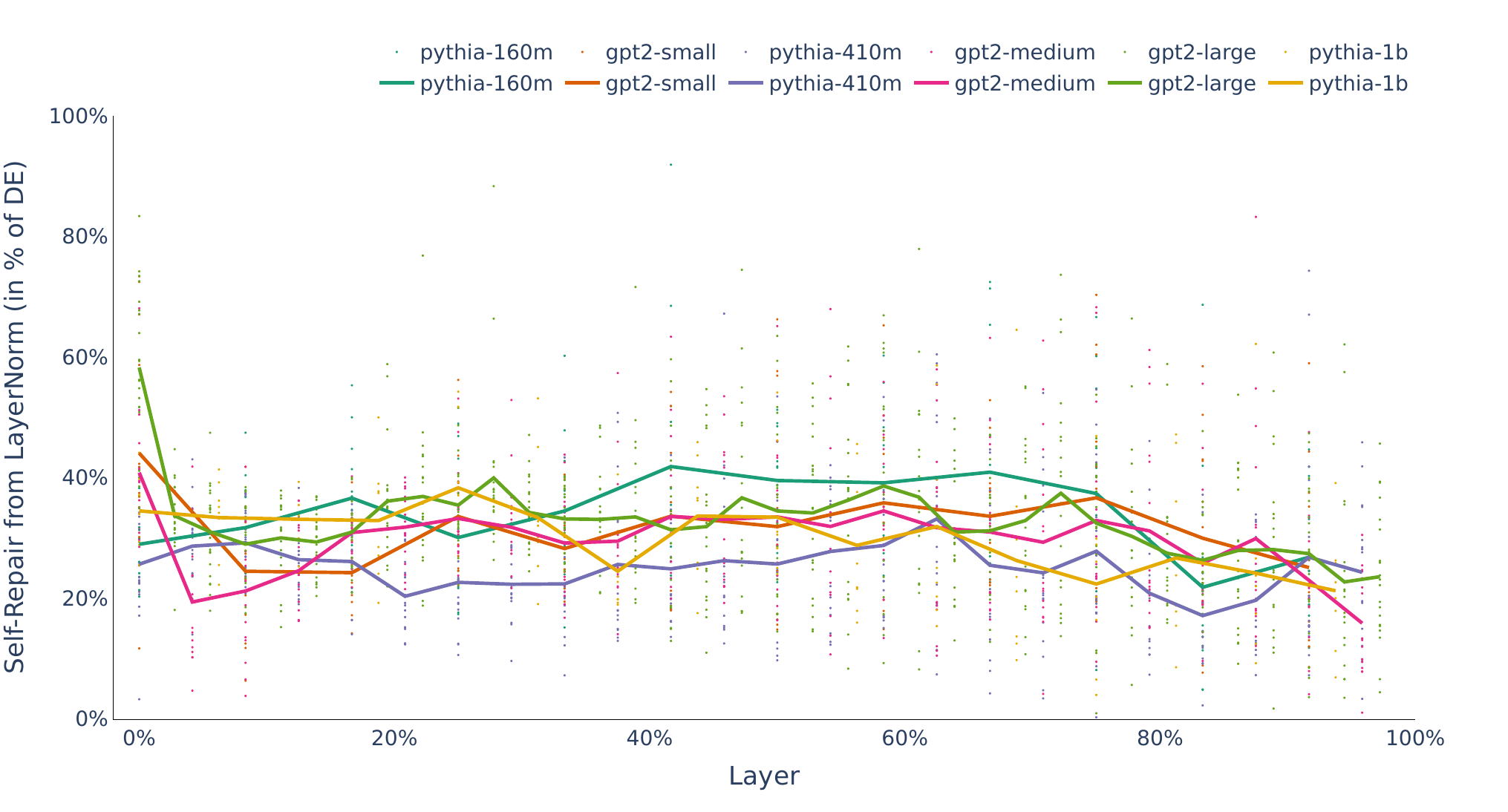}
\caption{Post-processed measuring of LayerNorm as a fraction of direct effect}
\label{fig:post-processed-layernorm-percentage}
\end{minipage}
\end{figure}

As an example, Figure \ref{fig:extreme-values} replicates the experiment in Section \ref{empirical-layernorm} where we quantify the amount of self-repair LayerNorm explains, but instead purely as a fraction of the total direct effect per head. Here we average across the entire distribution and do not clip the percentages for each token between 0 and 1. It's clear the values are extremely obtuse.

Thus, we attempt to create a cleaner representation of this result: our general two techniques for trying to capture more meaningful summary statistics were to 1) filter for the top instances of direct effect 2) clip the percentages of each token. These are nontrivial changes, and they have important impacts on the final figures we achieve. 

A full figure with these changes in effect is shown in Figure \ref{fig:post-processed-layernorm-percentage}, which is where we get our observation that "\LN of direct effect is self-repaired by LayerNorm": most of the models, on many of the layers, have an average self-repair due to LayerNorm which center around 30\%. The clipping on the values partially means that our figure for "\LN of direct effect is self-repaired by LayerNorm" can potentially be interpreted as "in \LN of cases, the self-repair due to LayerNorm is larger in magnitude than the direct effect of the head", given that the self-repair from LayerNorm, on many instances, is larger in magnitude than the actual direct effect of the head.

The noisiness of self-repair also means that even averaging across a layer hides a significant amount of nuance: the various attention heads have vastly different values. As such, we've plotted the individual heads in many of the graphs. However, summary statistics do not capture this nuance well.

\section{Mathematical Derivation of LayerNorm Self-Repair}
\label{derive-layernorm}

We show the derivation of LayerNorm's self-repair in full. Consider the simplified case where we analyze a parallel attention model such that ablating an attention head in the last layer only has a direct effect on the final residual stream. For an arbitrary model which uses LayerNorm, the model passes in the final residual stream $resid$ into the LayerNorm $LN$, and the logit for the correct token $logit$ is calculated by dotting the result with a vector $L$ which corresponds to the unembedding direction of the logit.

$$logit = \langle LN(resid), L\rangle $$

Assume we resample ablate an attention head in the last layer of attention heads with direct effect $DE_{\text{Head}}$. Consider the direct change in the residual stream, such that this change directly feeds into the new final residual stream $resid'$ with no other intermediate effects. This new residual stream maps onto the new correct token logit $logit' = \langle LN(resid'), L\rangle $.  If the change in the output of the head is $\Delta H$, then:

$$resid' = resid + \Delta H$$

Originally, one may predict that the change in logits $\Delta logit = logit' - logit$ is equal to $-DE_{\text{Head}}$, which models the ablation occuring in the absense of self-repair. Self-Repair is the observation that this isn't the case, and that often $|\Delta logit| < |DE_{\text{Head}}|$. 

How can we explain this logit difference? As argued in \citealt{elhage2021mathematical}{}, we can simplify the LayerNorm operation such that we `fold in' LayerNorm projections to the weights of linear layers before and after the projection, leaving the nonlinear operation of scaling the residual stream by dividing by a scaling factor proportional to the norm of the residual stream.

The two scaling factors on the clean and ablated run are $S$ and $S'$, which completely describe the LayerNorm functionality. With this, we can model what makes up the difference between the two logits: 
$$ \Delta logit = logit' - logit = \langle\frac{resid'}{S'}, L\rangle  - logit$$
Let's declare the change between $resid' - resid = \Delta H$, which is the difference in the output of head H as a result of ablating the attention head. As such, 

$$\Delta logit= (\langle \frac{resid}{S'}, L\rangle  + \langle \frac{\Delta H}{S'}, L\rangle ) - logit$$

And thus,

$$\Delta logit = (\frac{S}{S'})\langle \frac{resid}{S}, L\rangle  + \langle \frac{\Delta H}{S'}, L\rangle  - logit = (\frac{S}{S'})logit + \langle \frac{\Delta H}{S'}, L\rangle  - logit$$
The difference between the two residual streams is only the difference between the outputs between the clean and resample ablated head, $\Delta H =  H_{new} - H_{old}$. If you define the new direct effect of the resample ablated head as $DE_{\text{Head}}' = \langle \frac{H_{new}}{S}, L\rangle $, you can rewrite the above as
$$\Delta logit= (\frac{S}{S'} - 1)logit + (\frac{S}{S'})(\langle \frac{H_{new}}{S}, L\rangle  - \langle \frac{H_{old}}{S}, L\rangle )$$

And thus,
\[
\Delta \text{logit} = \underbrace{(\frac{S}{S'} - 1)\text{logit}}_{\text{LN on \textit{existing logits}}}  + \underbrace{\left(\frac{S}{S'}\right)(DE_{\text{Head}}' - DE_{\text{Head}})}_{\text{LN on Expected Change} \Delta DE_{\text{Head}}}
\]

% \section{Pythia-160m Newline Self-Repair}
% \label{newline-repair}

% L1H8 in Pythia-160m is has some strange functionality. When investigating how it was potentially Self-Repaired, we plotted its direct effect against the change in logits for different tokens in Figure \ref{fig:weird-pythia-self-repair}.

% % \begin{figure}
%     \centering
%     \includegraphics[width=0.5\linewidth]{weird_pythia_self_repair.png}
%     \caption{Ablating L1H8 induces strange model improvment. The top right cluster of tokens are newline tokens.}
%     \label{fig:weird-pythia-self-repair}
% \end{figure}

% There is this strange cluster of prompts at the upper right; these are instances where L1H8 contributes ~0.7 logits to the correct token, but when you ablate L1H8, model performance increases. Further analysis showed that this these changes were largely due to MLP layers downstream of the ablation.

% %Our current hypothesis for the behavior centers around 
% % > direct effect of MLP 11 when using clean cache as final LN scaling: 76.8090
% % > direct effect of MLP 11 when using clean cache as final LN scaling: 16.0072

% When investigating these, they are all instances where the Attention Head sees a newline and decides to (correctly) predict another newline. 

\section{Changes in MLP Neurons}
\label{appendix:changes-mlp-neurons}

\begin{figure}[h]
\centering
\begin{minipage}{0.47\linewidth}
\centering
\includegraphics[width=\linewidth]{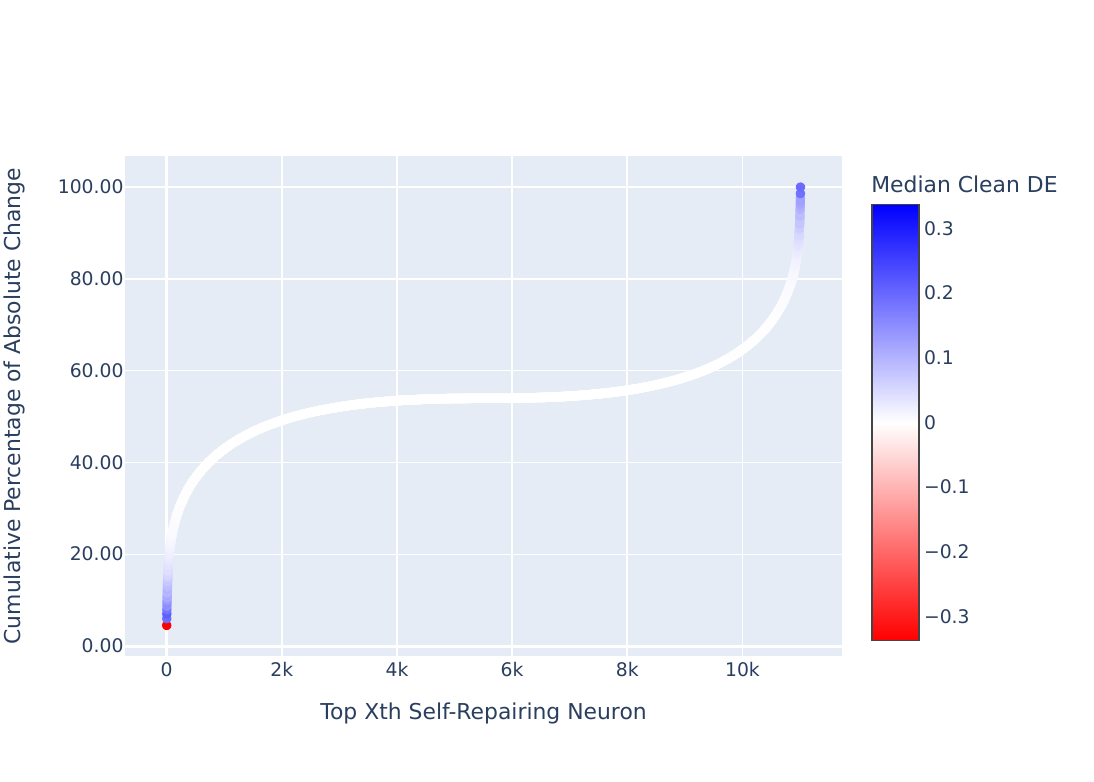}
\caption{Cumulative Percentage of Absolute Change in Direct Effect of neurons in the last layer of Llama-7B when ablating L30H8.}
\label{fig:llama-cumu-mlp}
\end{minipage}
\hfill
\vline
\hfill
\begin{minipage}{0.47\linewidth}
\centering
\includegraphics[width=\linewidth]{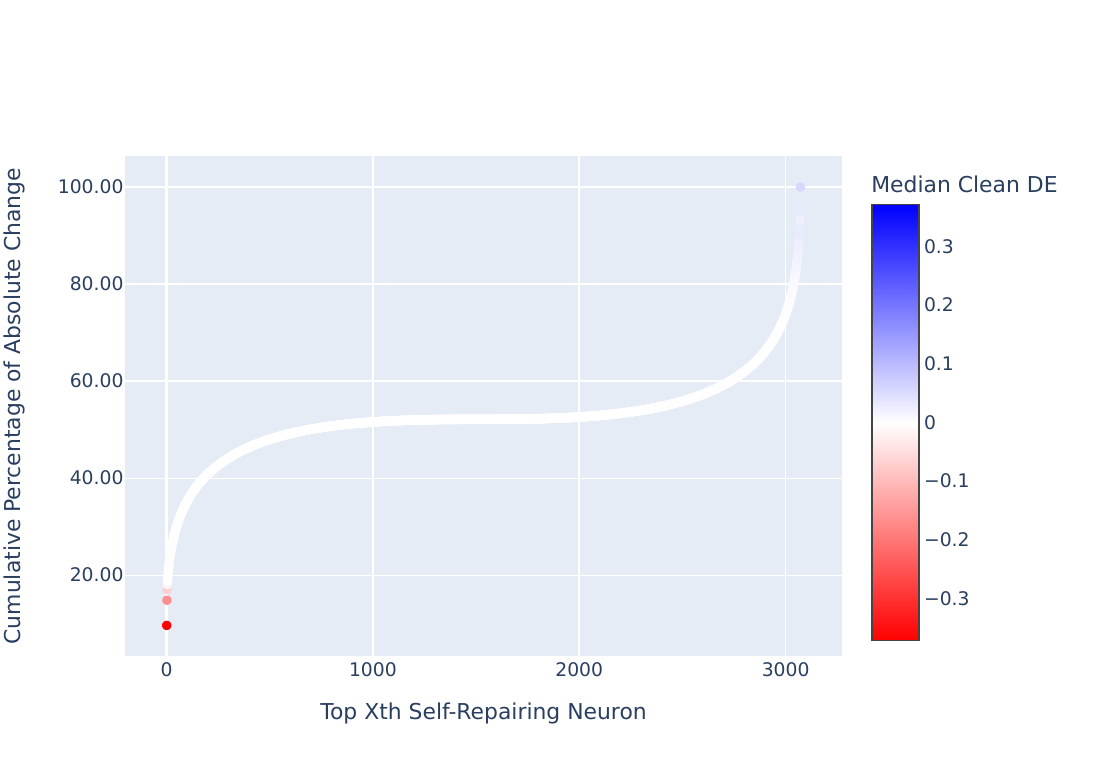}
\caption{Cumulative Percentage of Absolute Change in Direct Effect of neurons in the last layer of GPT2-Small when ablating L9H11.}
\end{minipage}

\medskip

\begin{minipage}{0.47\linewidth}
\centering
\includegraphics[width=\linewidth]{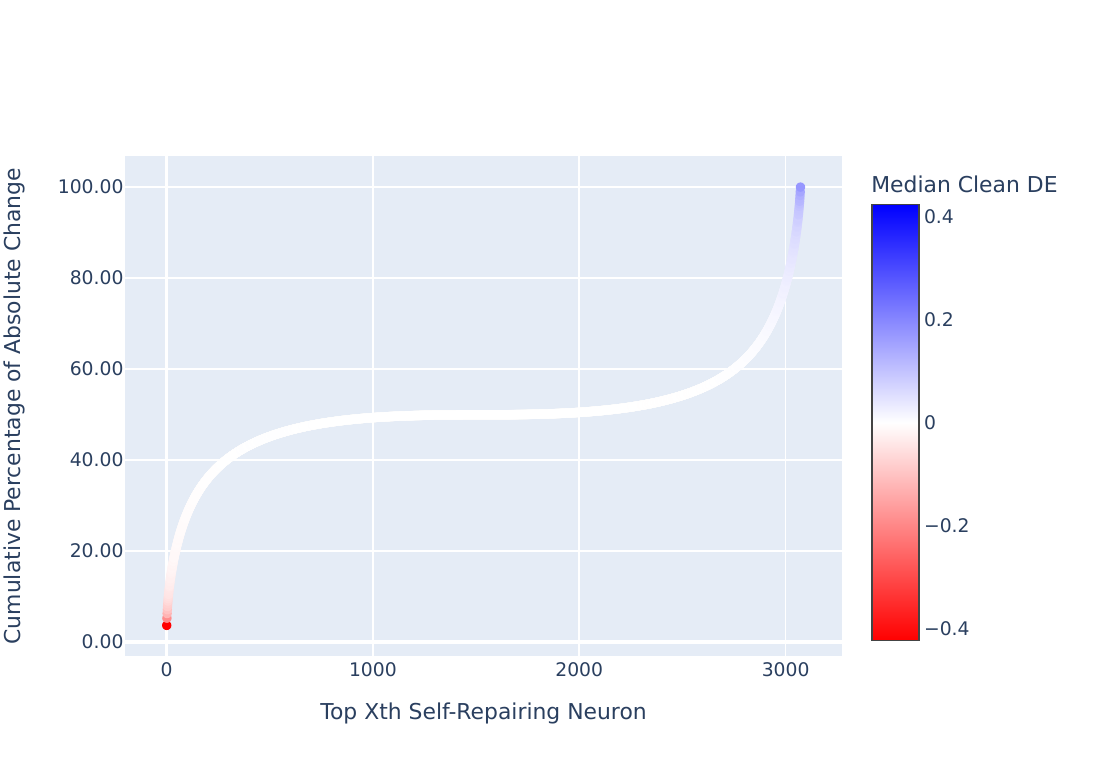}
\caption{Pythia-160M, Cumulative Percentage of Absolute Change in Direct Effect of neurons in the last layer when ablating L7H8.}
\end{minipage}
\hfill
\vline
\hfill
\begin{minipage}{0.47\linewidth}
\centering
\includegraphics[width=\linewidth]{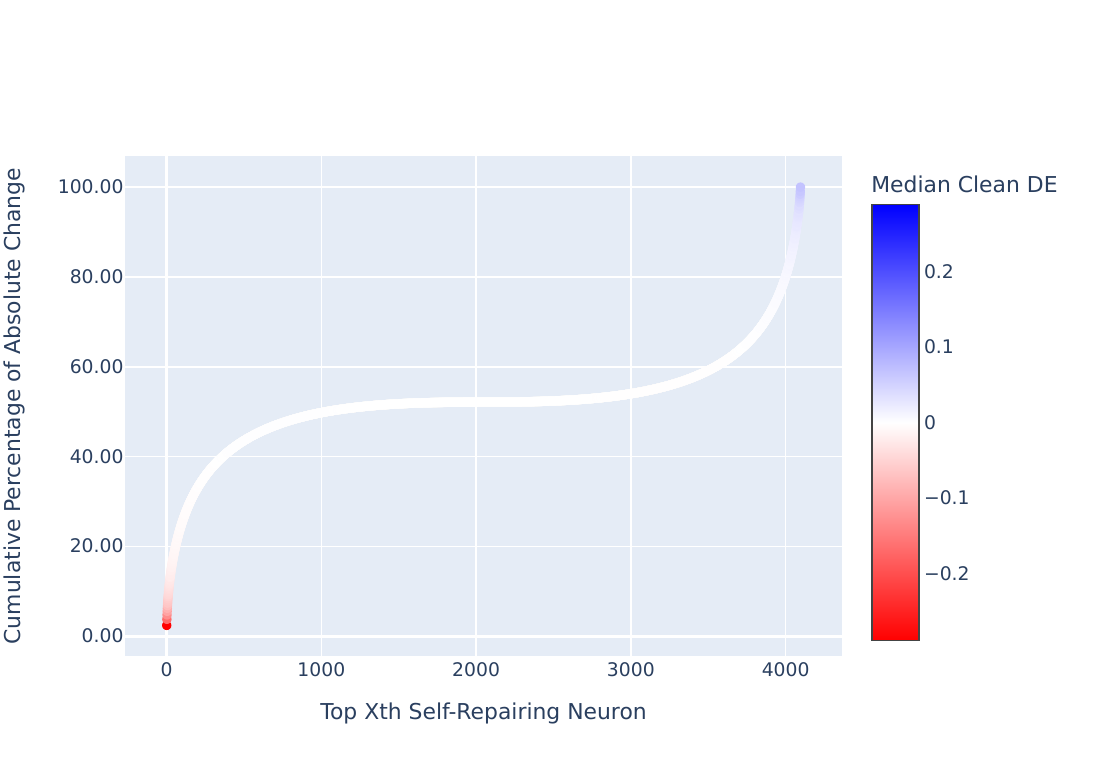}
\caption{Pythia-410M, Cumulative Percentage of Absolute Change in Direct Effect of neurons in the last layer when ablating L17H4.}
\label{fig:pythia-cumu-mlp}
\end{minipage}
\end{figure}

To produce Figures \ref{fig:llama-cumu-mlp} through \ref{fig:pythia-cumu-mlp}, we perform the same experimental setup as for Figure \ref{three_neurons_needed}, measuring the self-repair in the neurons of the final MLP layer upon the ablation of a specific attention head. However, in this graph, we have instead measured the absolute change in Direct Effect for each neuron. We sum all the absolute changes in direct effects and then plot the total cumulative percentage of absolute changes in direct effect explained up to the Top Xth Self-Repairing Neuron. For example, in Figure \ref{fig:pythia-cumu-mlp}, the sum of the individual absolute changes in direct effect of the top 1000 self-repairing neurons account for around 50\% of the total absolute changes in direct effect across the entire final MLP layer.

For each of these four graphs, it appears to be the case that roughly 60\% of the neurons in the layer have negligible changes in their absolute change: for instance, the neurons from the top 2000 to the top 9000 neurons in Figure \ref{fig:llama-cumu-mlp} explain less than 15\% of the total cumulative change in self-repair, and thus are negligible.

However, the other 40\% of neurons have some change in their direct effect. This implies that there may be other mechanisms for self-repair that cause smaller changes in direct effect across many neurons.

We've additionally colored each neuron with its median clean direct effect on the distribution. In almost all these graphs, the top repairing neurons have negative clean direct effects, and are thus erasure neurons (except for in Llama-7b, where only the top neuron is an erasure neuron on average).

Altogether, these support the claim that Sparse Anti-Erasure is one mechanism that helps self-repair in the model, but highlights how there are potentially more mechanisms besides it in MLP layers.

\section{Residual Stream Norms Change when Ablating}
\label{appendix:ablations-change-norms}

\begin{figure}[ht]
%\vskip 0.2in
\begin{center}
\centerline{\includegraphics[width=0.4\columnwidth]{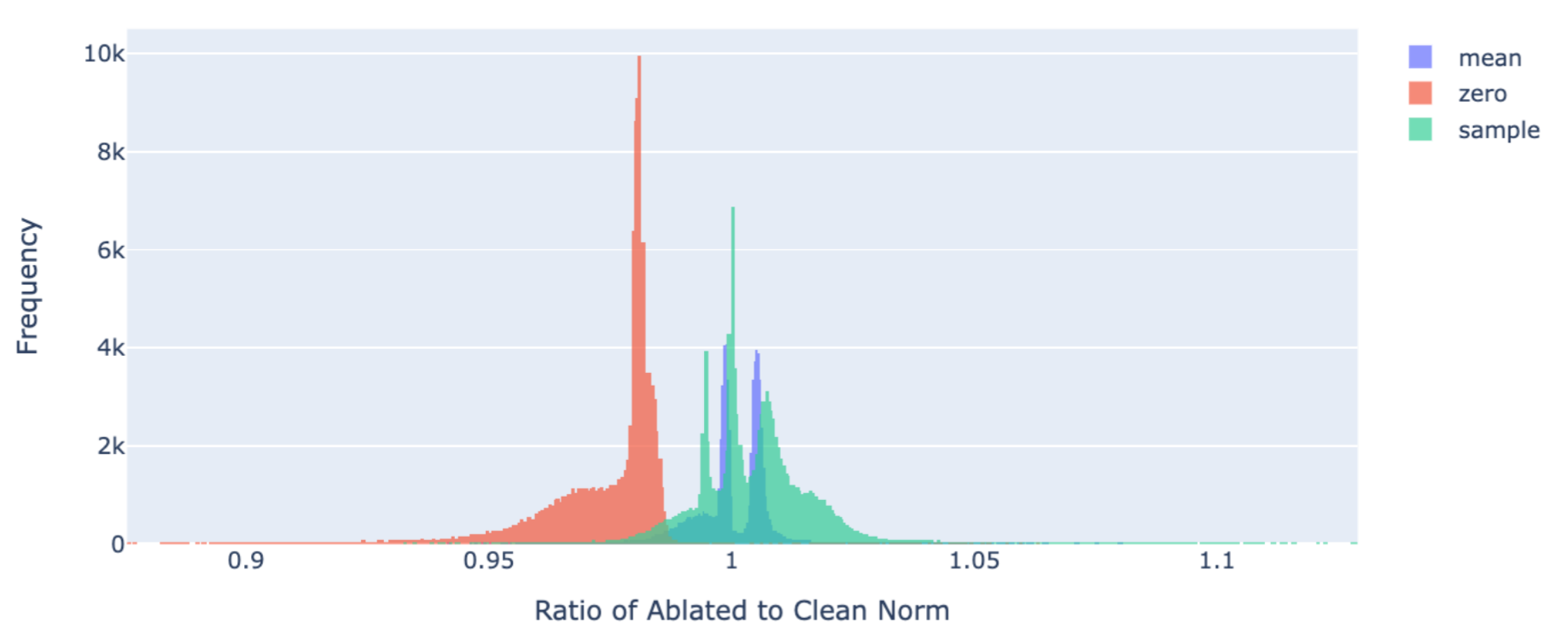}}
\caption{Ratio of ablated to clean residual stream norms when ablating L11H0 of Pythia-160m with various types of ablations, on The Pile.}
\label{different-ablation-LN}
\end{center}
%\vskip -0.2in
\end{figure}

Indeed, the residual stream norm changes most significantly as a result of zero ablations. In Figure \ref{different-ablation-LN}, we plot the ratio of ablated to clean residual stream norm for different ablations. Even without filtering for different direct effects, you can clearly see how zero ablating decreases the direct effect of a single head.

It's not surprising that a head in the last layer can decrease the residual stream norm by around $5\%$ when ablated. Past work found that residual stream norms grown exponentially \citep{exponential_growth}, so we would expect heads in later layers to be important for increasing norm.

\begin{figure}[h]
\centering
\begin{minipage}{0.47\linewidth}
\centering
\includegraphics[width=\linewidth]{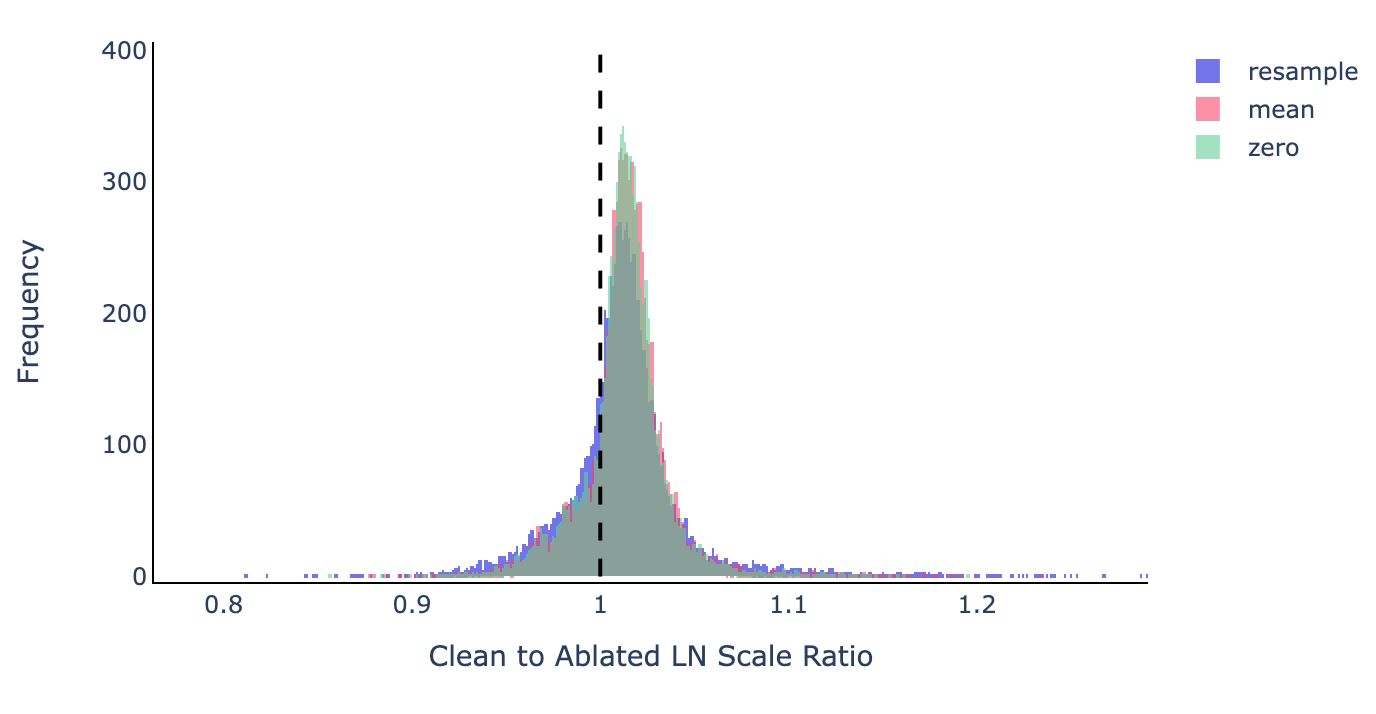}
\caption{LayerNorm scaling changes from ablating GPT2-Small L11H10. We follow the same experimental procedure outlined in Figure \ref{fig:ratio-LN-scales}.}
\label{fig:left-ablation-variance}
\end{minipage}
\hfill
\vline
\hfill
\begin{minipage}{0.47\linewidth}
\centering
\includegraphics[width=\linewidth]{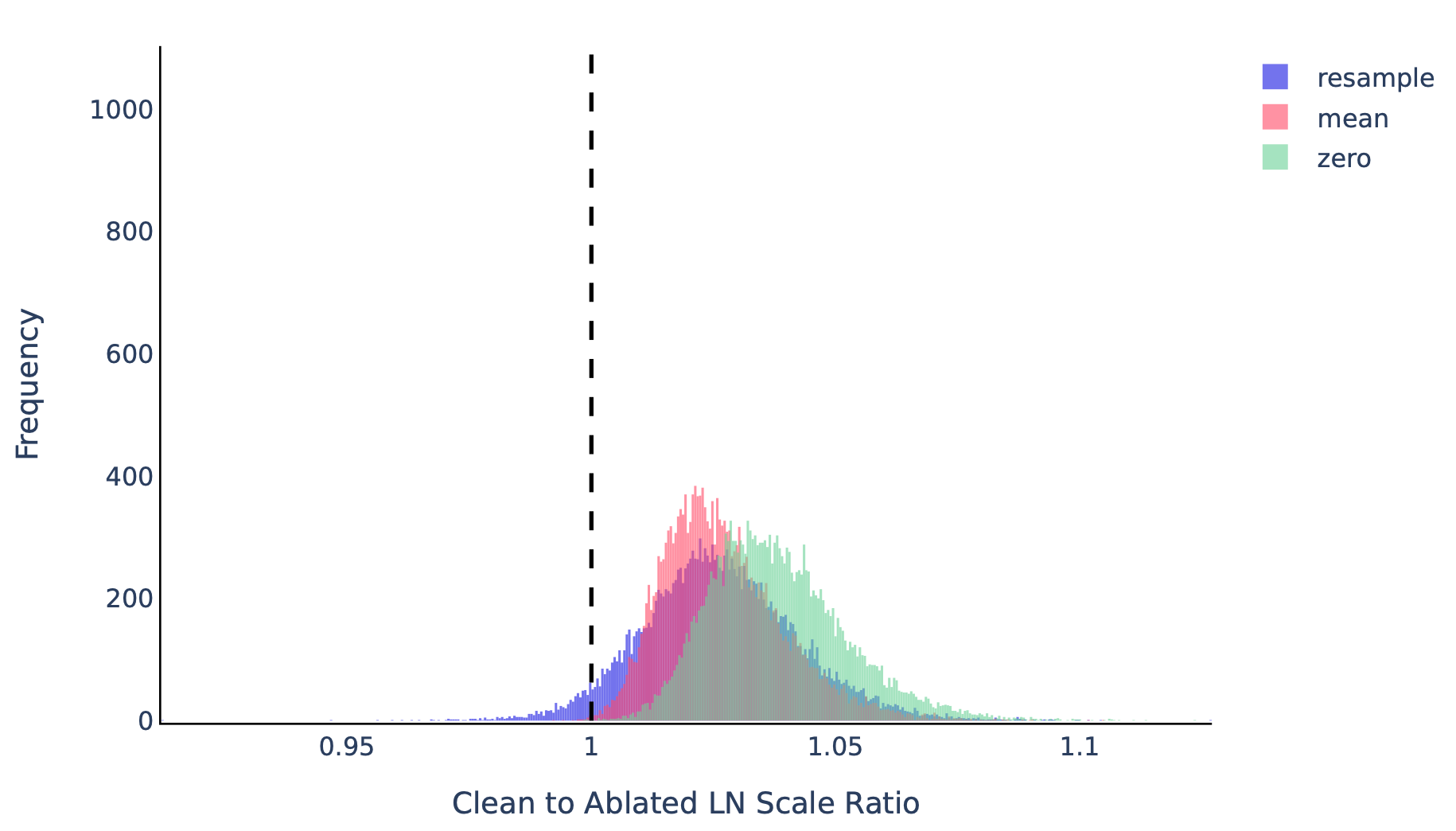}
\caption{LayerNorm scaling changes from ablating GPT2-Small L11H3. We follow the same experimental procedure outlined in Figure \ref{fig:ratio-LN-scales}.}
\label{fig:right-ablation-variance}
\end{minipage}
\end{figure}

For some heads, zero ablations induce more extreme changes on the LayerNorm: see additionally L11H3 of GPT2-Small, plotted in Figure \ref{fig:right-ablation-variance}. However, this does not always hold. For instance, all the different ablations seem to induce similar amounts of LayerNorm scaling ratios in L11H11 of GPT2-Small, plotted in Figure \ref{fig:left-ablation-variance}.

\section{Centered Logits}
\label{center-logits}
An important aspect of our analysis is that the logits at each position are centered such that the average logit is zero. This helps control for changes in logits which can be explained by \textit{translations in all the logits} - which is important, because a translation in logits leads to no change in the final log probabilities of the outputs.

As adding a constant to the logits does not affect the output probabilities (softmax is translation invariant), centered logits are more amenable to analysis. As such, we cannot measure parts of self-repair coming from adding a constant to the logits of all tokens simultaneously, as the centering of the logits would remove this behavior. Instead, the observation of self-repair in a logit - such as in Figure \ref{ablation-comparison} - indeed reflects a self-repair in the correct token (and not all tokens).

\section{Self-Repressing Heads}
\label{appendix:self-repress-scaling}

\begin{figure}
    \centering
    \includegraphics[width=0.5\linewidth]{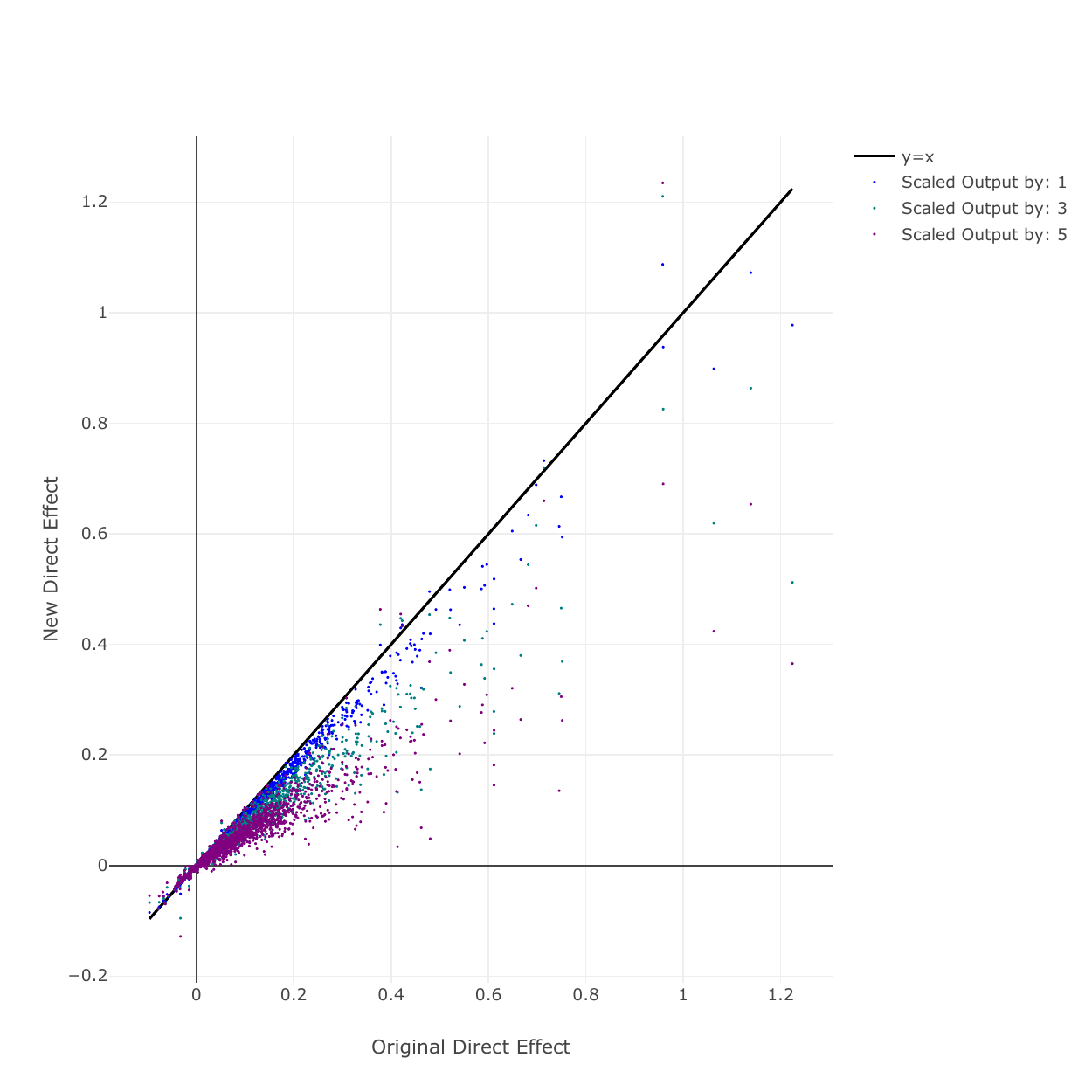}
    \caption{Self-Repressing Head L21H1 in GPT2-Medium being self-repressed by various amounts when amplifying the input into the residual stream by various amounts.}
    \label{fig:self-repress-scales}
\end{figure}

The existence of Self-Repressing heads (Section \ref{sec:model-iterativity}) helps support the Iterative Inference hypothesis because they robustly respond to their own output. However, one interesting aspect of this is that the amount in which these heads self-repress is proportional to the amount in which scale the heads output back into itself. Figure \ref{fig:self-repress-scales} highlights the self-repressing head L21H1 in GPT2-Medium, but plots the forward passes when adding in the output of the head scaled by 1, 3, and 5 times the existing output.

\section{LayerNorm Ratios are Significant}
\label{app:ln-ratios}
For many models, the direct effect of the head is often ‘significant’ when its direct effect is greater than 1 (it can get to up to 4, but on rare occasions). However, the difference between two logits can be up to around 10 or more logits! For a ratio such as 1.02, a difference of 10 logits between two tokens would lead to a difference of 0.2 logits in how much LayerNorm self-repairs one token over another in the toy example presented in Section \ref{section_layernorm} (the amount a token A is self-repaired relative to another token B can be calculated via $(\frac{S}{S’} - 1)(logit_a - logit_b)$, where $logit_a$ and $logit_b$ are the final logits of A and B - this equation can be easily derived from the first presented equation in Section \ref{section_layernorm}). This is a nontrivial fraction of a direct effect.

\section{Self-Repair Graphs Across Different Models.}
\label{across-models-graphs}
\begin{figure}[h]
\centering
\begin{minipage}{0.47\linewidth}
\centering
\includegraphics[width=\linewidth]{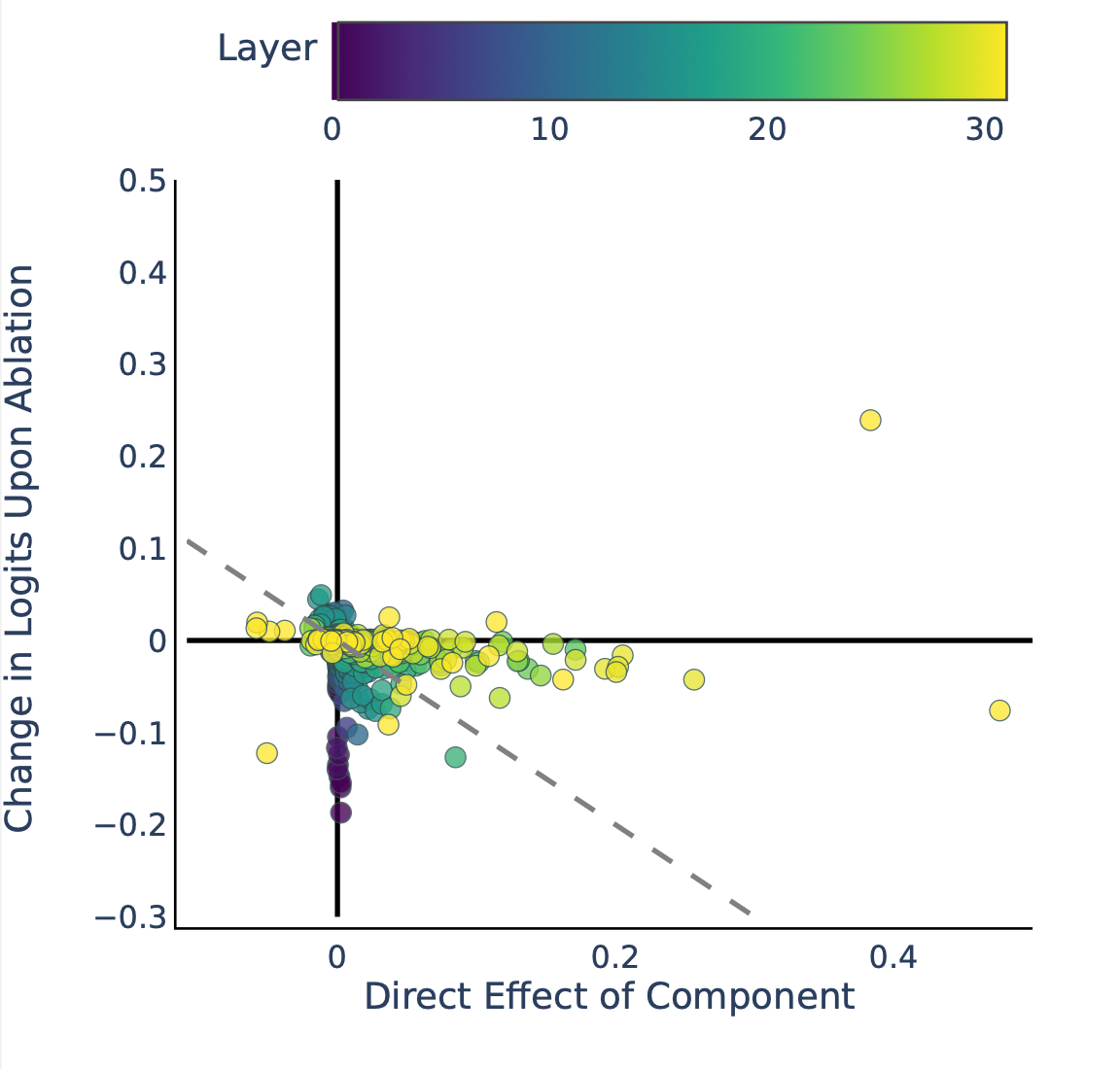}
\caption{Llama-7b self-repair, per head}
\label{fig:start-of-graph}
\end{minipage}
\hfill
\vline
\hfill
\begin{minipage}{0.47\linewidth}
\centering
\includegraphics[width=\linewidth]{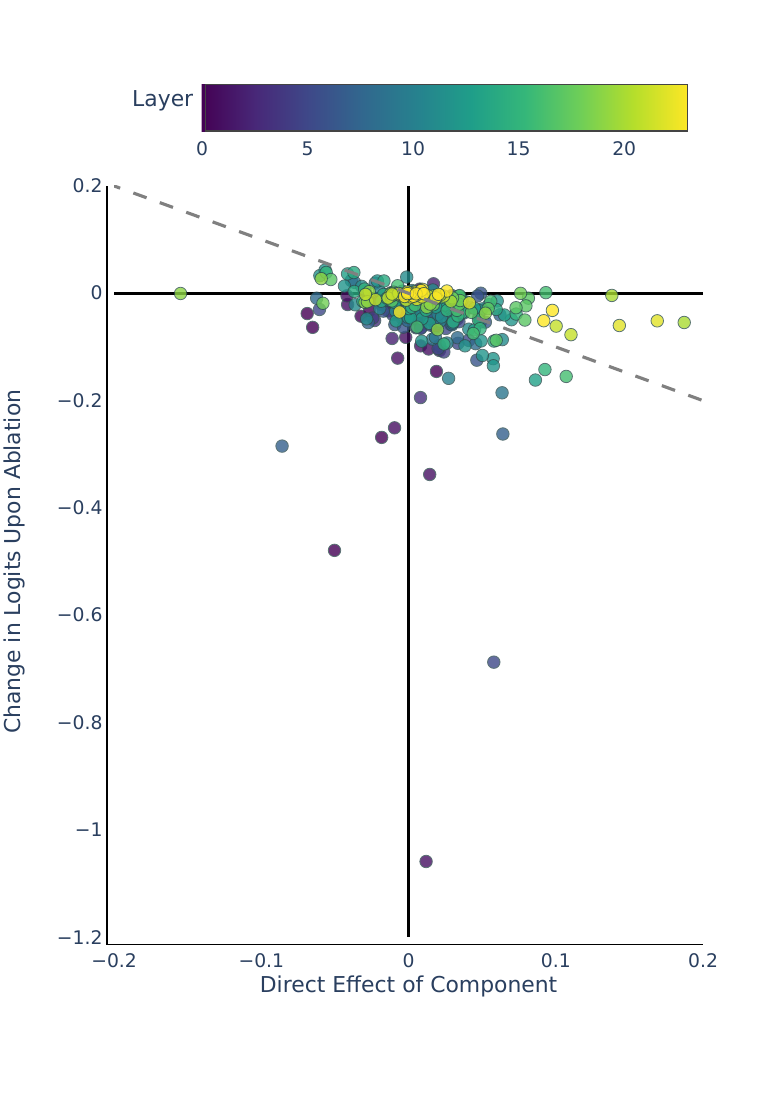}
\caption{Pythia-410m self-repair, per head}
\end{minipage}
\end{figure}

\begin{figure}
    \centering
    \includegraphics[width=0.5\linewidth]{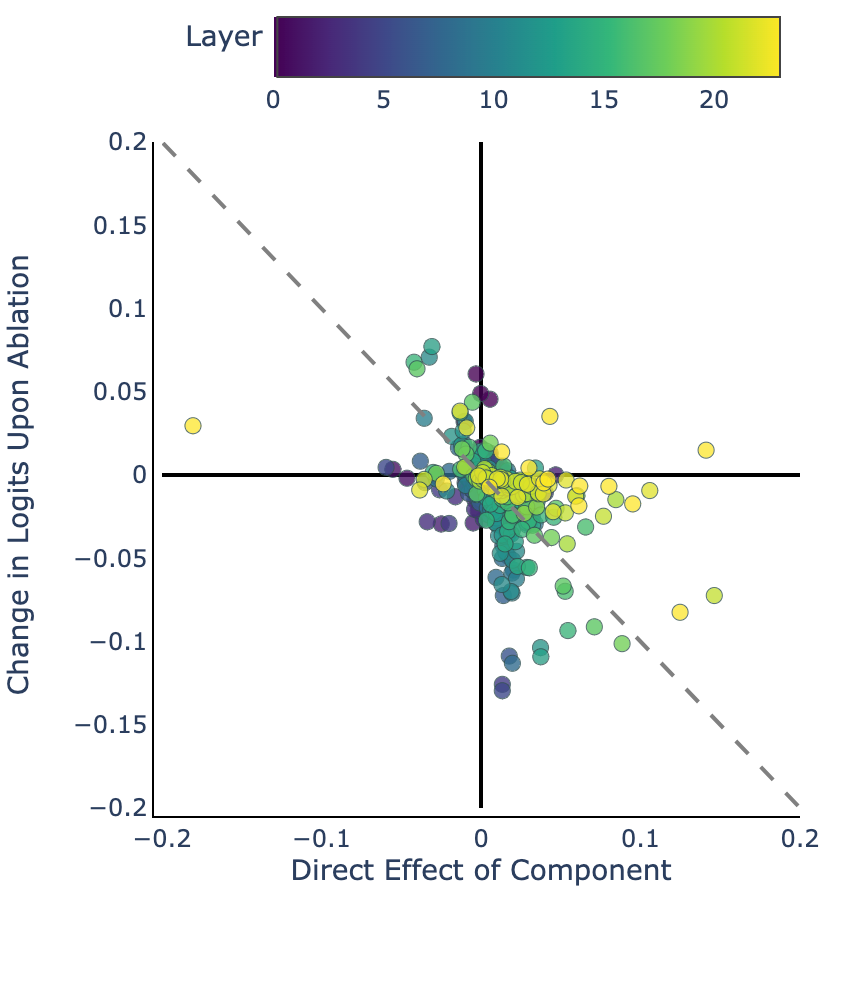}
    \caption{GPT2-Medium self-repair, per head}
    \label{fig:end-of-graph}
\end{figure}

\begin{figure}[h]
\centering
\begin{minipage}{0.47\linewidth}
\centering
\includegraphics[width=\linewidth]{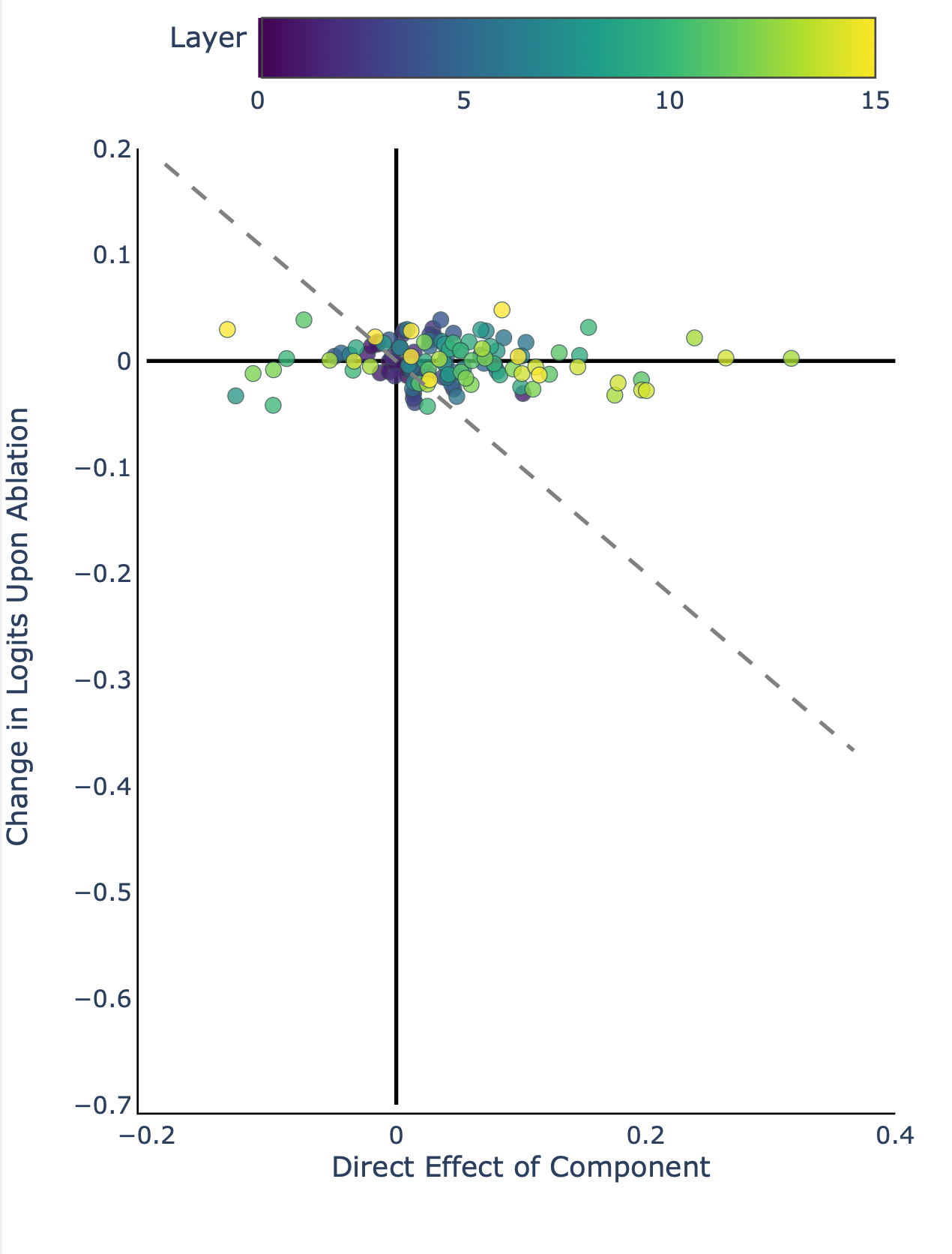}
\caption{An idealized version of perfect self-repair}
\label{fig:perfect-sr}
\end{minipage}
\hfill
\vline
\hfill
\begin{minipage}{0.47\linewidth}
\centering
\includegraphics[width=\linewidth]{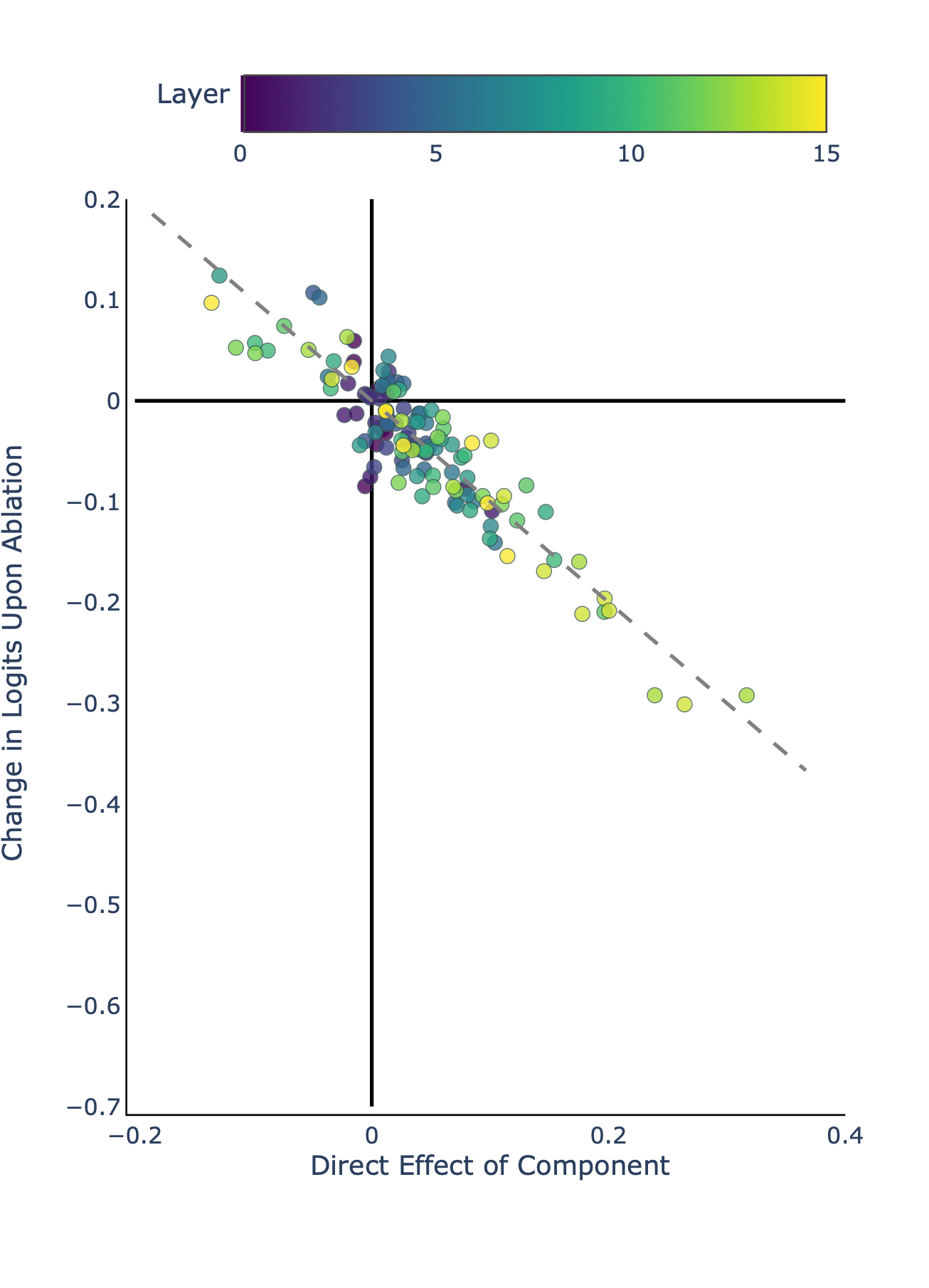}
\caption{An idealized version of no self-repair}
\label{fig:no-sr}
\end{minipage}
\end{figure}

In Figures \ref{fig:start-of-graph} through \ref{fig:end-of-graph}, we plot the self-repair of individual attention heads for different models, similar to the experimental setup for Figure \ref{fig:self-repair-scatter-main}.  We also highlight what idealized versions of `perfect self-repair' and no self-repair my look like in Figures \ref{fig:perfect-sr} and \ref{fig:no-sr}.

\end{document}